%% file: lightest.tex
\documentclass[runningheads]{llncs}
\usepackage{graphicx}
\usepackage{comment}
\usepackage{amsmath,amssymb} 
\usepackage{color}

\usepackage{booktabs} 
\usepackage{array}
\usepackage{makecell}

\begin{document}
\pagestyle{headings}
\mainmatter
\def\ECCVSubNumber{2353}  

\title{Object-based Illumination Estimation with Rendering-aware Neural Networks} 

\titlerunning{Rendering-aware Illumination Estimation}
%
\author{Xin Wei\inst{1,2} \and
Guojun Chen\inst{1} \and
Yue Dong \inst{1} \and
Stephen Lin \inst{1} \and
Xin Tong \inst{1}}
\authorrunning{X. Wei et al.}
%
\institute{Microsoft Research Asia \and 
Zhejiang Univiersity}
\maketitle

\graphicspath{{fig/}}
\input{cmd_def}

\input{abstract}

\input{intro}
\input{related}
\input{overview}

\input{network}
\input{training}
\input{results}

\input{conclusion}

\clearpage
%
%
\bibliographystyle{splncs04}
\bibliography{fastlightest}
\end{document}


\title{Object-based Illumination Estimation with Rendering-aware Neural Networks --Supplemental Material--}

\renewcommand{\shortauthors}{}
\graphicspath{{fig/}}
\input{cmd_def}

\input{fig_teaser}

\maketitle

\section{Additional real examples}
In this supplementary material, we include additional real-world measured examples, inserting virtual objects into the image. We also compared with the composition rendered with measured ground truth environment light. Figure ?? to ?? illustrates 10 such examples, and the average lighting and render RMSE are ?? and ??. 

\section{Additional synthetic examples}
We include the complete test results on our synthetic test dataset, shown from Figure ?? to ??, with numerical error for each individual examples provided. 
We also show selected synthetic example that shows all the intermediate results from each individual network as well as the output of the intermediate supervision. 

\section{Implementation details}

\paragraph{Training dataset}
Figure ?? shows a selective set of examples from our training dataset. 

\paragraph{Network structures}
Figure ?? illustrates the detailed network structure of our system. 



\bibliographystyle{ACM-Reference-Format}
\bibliography{fastlightest}

%% file: cmd_def.tex

\definecolor{dblue}{rgb}{0.0,0.0,0.5}
\definecolor{dgreen}{rgb}{0.0,0.5,0.0}
\definecolor{dred}{rgb}{0.6,0.3,0.0}
\definecolor{dorange}{rgb}{0.6,0.25,0.0}
\definecolor{dyellow}{rgb}{0.5,0.5,0.0}
\newcommand{\NOTE}[1]{ \texttt{ \textbf{NOTE:} #1~ } }
\newcommand{\TODO}[1]{ \textcolor{blue}{\texttt{ \textbf{TODO:} #1~ }} }
\newcommand{\Yue}[1]{ \textcolor{dorange}{ (\emph{Yue}: #1)}  }

\newcommand{\intd}{\,\text{d}}

\newcommand{\InputRGB}{\mathbf{I}}
\newcommand{\InputDepth}{\mathbf{D}}
\newcommand{\Light}{\mathbf{L}}
\newcommand{\Irr}{\mathbf{E}}
\newcommand{\Vis}{\mathbf{V}}
\newcommand{\Normal}{\mathbf{N}}
\newcommand{\TransMat}{\mathbf{T}}
\newcommand{\Mat}[1]{\mathbf{#1}}

\newcommand{\mycaptionbox}[1]{\begin{minipage}{\textwidth}\vspace*{-1.0cm}\hspace*{0.1cm}\small{#1}\end{minipage}\vspace*{-1.0cm}}

\newcommand{\fiveColumnFigWidth}{1.2cm} 
\newcommand{\fourColumnFigWidth}{1.5cm}
\newcommand{\threeColumnFigWidth}{2.95cm}

\newcommand{\realInputSize}{2cm}
\newcommand{\realRenderSize}{5cm}

%% file: abstract.tex
\begin{abstract}

We present a scheme for fast environment light estimation from the RGBD appearance of individual objects and their local image areas. Conventional inverse rendering is too computationally demanding for real-time applications, and the performance of purely learning-based techniques may be limited by the meager input data available from individual objects. To address these issues, we propose an approach that takes advantage of physical principles from inverse rendering to constrain the solution, while also utilizing neural networks to expedite the more computationally expensive portions of its processing, to increase robustness to noisy input data as well as to improve temporal and spatial stability. This results in a rendering-aware system that estimates the local illumination distribution at an object with high accuracy and in real time. With the estimated lighting, virtual objects can be rendered in AR scenarios with shading that is consistent to the real scene, leading to improved realism.

\end{abstract}

%% file: intro.tex
\section{Introduction}
\label{sec:intro}

Consistent shading between a virtual object and its real-world surroundings is an essential element of realistic AR. To achieve this consistency, the illumination environment of the real scene needs to be estimated and used in rendering the virtual object. For practical purposes, the lighting estimation needs to be performed in real time, so that AR applications can accommodate changing illumination conditions that result from scene dynamics.

Traditionally, lighting estimation has been treated as an inverse rendering problem, where the illumination is inferred with respect to geometric and reflectance properties of the scene \cite{Sato2003IFS,Romeiro:2010:BR,Barron2015,Lombardi2016a}. Solve the inverse rendering problem with a single image input is ill-conditioned, thus assumptions are made to simplify the lighting model or to limit the supported material type. Despite the ambiguity among shape, material, and lighting. Solving the optimization involved in inverse rendering entails a high computational cost that prevents real-time processing, 
since the forward rendering as a sub-step of such optimization already difficult to reach real-time performance without compromising the accuracy.

Recent methods based on end-to-end neural networks provide real-time performance \cite{holdgeoffroy2017,Gardner2017,Cheng2018,LeGendre:2019:DLI,Gardner_2019_ICCV,Song_2019_CVPR}, specifically, they regard the input image as containing partial content of the environment map and estimate high resolution environment light based on those contents,
rich content in the input image is critical to infer the surrounding environment without ambiguity. However, many AR applications have interactions focused on a single object or a local region of a scene, where the partial content of the environment map in the input image is very limited. Thinking about guessing an environment map based on the image of a toy putting on the ground (like Figure \ref{fig:light_size_num}.h), although the ground is part of the environment map, it provides limited clues to rule out ambiguities when inferring the surroundings, yielding unstable estimation results. 

On the contrary, given such a scenario, the appearance of the object and the shadow cast by the object provide strong cues for determining the environment light. 
Utilizing such cues 	with neural networks is however challenging, due to the complex relationship between the lighting, material, and appearance. First, the neural network needs to aware of the physical rules of the rendering. 
Second, physically based rendering is computationally intensive, simple analysis by synthesis is not suitable for a real-time application.
Third, the diffuse and specular reflections follow different rules and the resulting appearance are mixed together in the input based on the unknown material property of the object. Previous methods already prove optimizing an end-to-end model for such a complex relationship is challenging and inefficient \cite{Song_2019_CVPR}. 

In this paper, we present a technique that can estimate illumination from the RGBD appearance of an individual object and its local image area. 
To make the most of the input data from a single object, our approach is to integrate physically-based principles of inverse rendering together with deep learning. An object with a range of known surface normals provides sufficient information for lighting estimation by inverse rendering. To deal with the computational inefficiency of inverse rendering, we employ neural networks to rapidly solve for certain intermediate steps that are slow to determine by optimization. Among these intermediate steps are the decomposition of object appearance into reflection components -- namely albedo, diffuse shading, and specular highlights -- and converting diffuse shading into a corresponding angular light distribution. On the other hand, steps such as projecting specular reflections to lighting directions can be efficiently computed based on physical laws without needing neural networks for speedup. Estimation results are obtained through a fusion of deep learning and physical reasoning, via a network that also accounts for temporal coherence of the lighting environment through the use of recurrent convolution layers.

In this way, our method takes advantage of physical knowledge to facilitate inference from limited input data, while making use of neural networks to achieve real-time performance. This rendering-aware approach moreover benefits from the robustness of neural networks to noisy input data. Our system is validated through experiments showing improved estimation accuracy from this use of inverse rendering concepts over a purely learning-based variant. The illumination estimation results also compare favorably to those of related techniques, and are demonstrated to bring high-quality AR effects. 

%% file: related.tex
\section{Related Work}
\label{sec:related}

A large amount of literature exists on illumination estimation in computer graphics and vision. Here, we focus on the most recent methods and refer readers to the survey by Kronander et al.~\cite{Kronander:2015:PRM} for a more comprehensive review.

\paragraph{Scene-based lighting estimation}
Several methods estimate lighting from a partial view of the scene. In earlier works, a portion of the environment map is obtained by projecting the viewable scene area onto it, and the rest of the map is approximated through copying of the scene area~\cite{Khan:2006:IME} or by searching a panorama database for an environment map that closely matches the projected scene area~\cite{Karsch:2014:ASI}.

Deep learning techniques have also been applied to this problem. Outdoor lighting estimation methods usually take advantage of the known prior distribution of the sky and predict a parametric sky model~\cite{holdgeoffroy2017,Zhang_2019_CVPR} or a sky model based on a learned latent space~\cite{Hold-Geoffroy_2019_CVPR}. Another commonly used low-parametric light model is the spherical harmonic (SH) model, adopted by \cite{Cheng2018,Garon_2019_CVPR}. However, an SH model is generally inadequate for representing high frequency light sources. Recently, Gardner et al.~\cite{Gardner_2019_ICCV} proposed a method estimating a 3D lighting model composed of up to 3-5 Gaussian lights and one global ambient term, improving the representation power of the parametric light model.
Instead of depending on a low-order parametric model, our method estimates high resolution environment maps without the limitations of such models.

Recent scene-based methods~\cite{Gardner2017,LeGendre:2019:DLI,Song_2019_CVPR} regard the input image as containing partial content of the environment map and estimate high resolution environment light based on those contents.
The input image is usually regarded as a warped partial environment map that follows the spherical warping \cite{Gardner2017} or warping based on depth \cite{Song_2019_CVPR}. The quality their estimated light depends on the amount of content in the known partial environment, since fewer content in the partial environment leads to stronger ambiguity of the missing part, increasing the estimation difficulties. By contrast, our work seeks to estimate an environment map from only the shading information of an object, without requiring any content of the environment map, which is orthogonal to scene-based methods and could be a good complement.

\paragraph{Object-based lighting estimation}
Illumination has alternatively been estimated from the appearance of individual objects in a scene. A common technique is to include a mirrored sphere in the image and reconstruct the lighting environment through inverse rendering of the mirrored reflections~\cite{Debevec:1998:RSO,Waese2002,Unger2006}. This approach has been employed with other highly reflective objects such as human eyes \cite{Nishino:2004:ER}, and extended to recover multispectral lighting using a combination of light probes and color checker charts \cite{LeGendre:2016:PML}.

Avoiding the use of special light probes, several inverse rendering methods utilize general objects instead. Some estimate lighting from diffuse objects for which rough geometry is measured by a depth sensor \cite{Gruber2012,Barron2013,Gruber2014}, reconstructed by multi-view stereo \cite{Wu2011}, or jointly estimated with the lighting~\cite{Barron2015}. For homogeneous objects with a known shape, the BRDF has been recovered together with all-frequency illumination~\cite{Romeiro:2010:BR,Lombardi2016a}. Cast shadows from objects with known geometry have also been analyzed to extract high-frequency lighting information~\cite{Sato2003IFS,Jiddi2017}. With a complete scan and segmentation of a scene, inverse path-tracing has been used to estimate both the light and object BRDFs~\cite{Azinovic_2019_CVPR}. Inverse rendering approaches such as these have proven to be effective, but are too computationally slow for real-time AR applications due to iterative optimization. Recently, Sengupta et al.~\cite{sengupta2019neural} trained a inverse rendering neural network; however, their direct rendering component assumes diffuse shading only and their method produces a low-resolution environment map limited by the direct renderer.

Deep learning has also been applied for illumination estimation from objects. Many of these works specifically address the case of human faces, for which prior knowledge about face shape and appearance can be incorporated \cite{Tewari:2017:MDC,Calian:2018:FFO,Yi:2018:FLP,Tewari:2018:SMF,Zhou_2018_CVPR,Sengupta_2018_CVPR,weber2018learning,Sun:2019:SIP}. For more general cases, neural networks have been designed to estimate an environment map from homogeneous specular objects of unknown shape~\cite{GeorgoulisPAMI2017} and for piecewise constant
materials while considering background appearance \cite{GeorgoulisICCV17}. By contrast, our approach does not make assumptions on the surface reflectance, and makes greater use of physical principles employed in inverse rendering.

%% file: overview.tex
\section{Overview}
\label{sec:overview}

\input{fig_overview}

Our method takes RGBD video frames from an AR sensor as input. For a cropped object area within a frame, an illumination environment map is estimated for the object's location. Illustrated in Fig.~\ref{fig:env}, the estimation process consists of three main components: reflectance decomposition, spatial-angular translation, and angular fusion.

{\em Reflectance decomposition} aims to separate object appearance into albedo, diffuse shading, and specular reflections. Such a decomposition facilitates inverse rendering, as diffuse and specular reflection arise from different physical mechanisms that provide separate cues about the lighting environment. 

{\em Spatial-angular translation} then converts the computed diffuse and specular shading maps from the spatial domain into corresponding lighting distributions in the angular domain. 
Since the relationship between lighting and shading is different for diffuse and specular reflections, we perform the spatial-angular translation of diffuse and specular shading separately.
For diffuse shading, a neural network is trained to infer a low-resolution environment map from it. On the other hand, the translation from the specular map to lighting angles is computed geometrically based on mirror reflection and a normal map of the object, calculated from its depth values. 

The angular lighting maps from diffuse and specular reflections are then merged through an {\em angular fusion} network to produce a coherent environment map based on the different reflectance cues, as detailed in Sec.~\ref{sec:fusion}. To ensure temporal consistency of the estimated environment maps over consecutive video frames, we incorporate recurrent convolution layers \cite{Huang2015,Chaitanya:2017:IRM} that account for feature maps from previous frames in determining the current frame's output.

%% file: fig_overview.tex

\begin{figure}[t]
   \centering\includegraphics[width=0.9\textwidth]{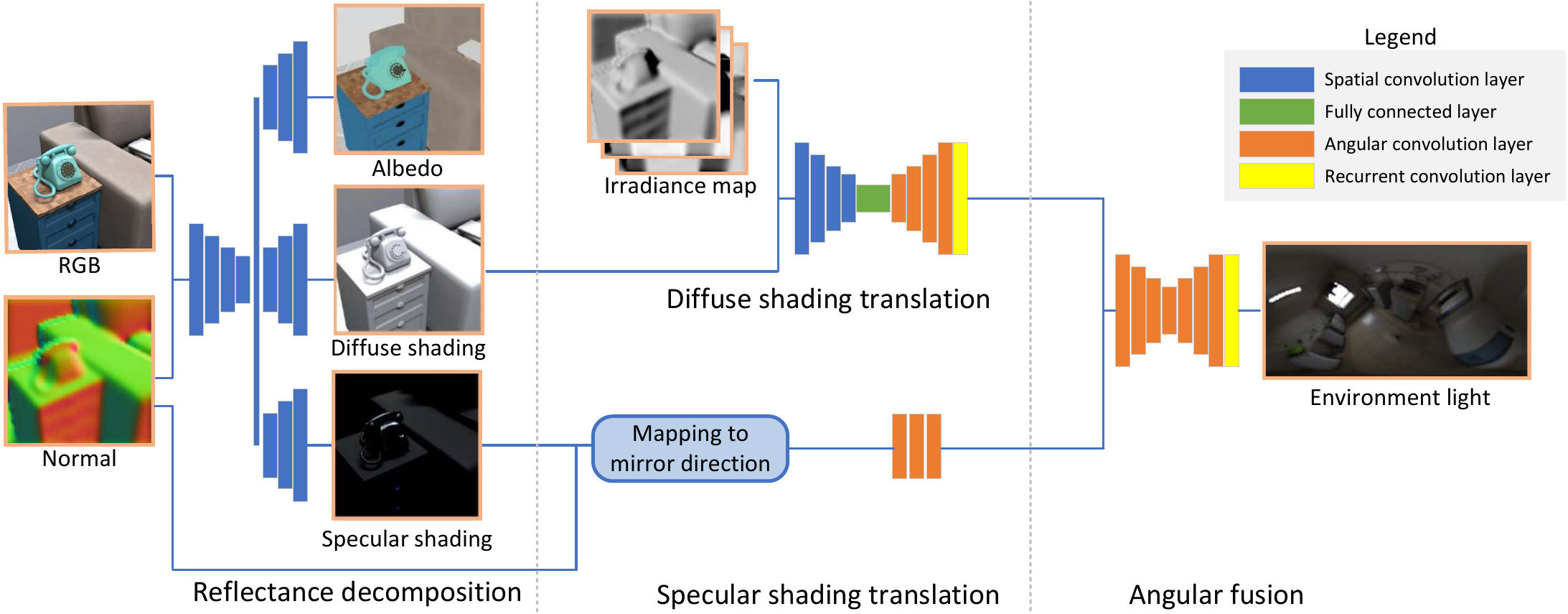}
    \caption{Overview of our system. The input RGB image crop is first decomposed into albedo, diffuse shading and specular shading maps. This decomposition is assisted by shape information from a normal map, calculated from rough depth data. The diffuse shading is translated into the angular lighting domain with the help of auxiliary irradiance maps computed from the rough depth, and the specular shading is geometrically mapped to their mirror direction. The translated features are then processed in the angular fusion network to generate the final environment light estimate.}
    \label{fig:overview} 
\end{figure}

%% file: network.tex
\vspace*{-0.1cm}
\section{Network structures}
\label{sec:network}

\paragraph{Reflectance decomposition network}
\label{sec:decomposition}
The input of the reflectance decomposition network is a cropped RGB input image and the corresponding normal map computed from the bilateral filtered input depth map as the cross-product of depths of neighboring pixels.
After a series of convolutional,
downsampling and upsampling layers, the network produces a set of decomposed maps including the diffuse albedo map, diffuse shading map, and specular shading map. Our reflectance decomposition is similar to traditional intrinsic image decomposition, but additionally separates shading into diffuse and specular components. Moreover, the normal map taken as input provides the network with shape information to use in decomposition, and the network also generates a refined normal map as output.

In practice, we follow the network structure and training loss of Li et al.~\cite{Li:2018:CGI} with the following modifications. We concatenate the normal map estimated from the rough depth input as an additional channel to the input.
We also split the shading branch into two sibling branches, one for diffuse shading and the other for specular shading, each with the same loss function used for the shading component in~\cite{Li:2018:CGI}.
The exact network structure is illustrated in the supplementary material.

\paragraph{Diffuse shading translation}
The relationship between the angular light distribution and the spatial diffuse shading map is non-local, as light from one angular direction will influence the entire image according to the rendering equation:
\begin{equation}
    R(x,y) = v(x,y) n(x,y) \cdot l
\end{equation}
where $v(x,y)$ denotes the visibility between the surface point at pixel $(x,y)$ and the light
direction $l$, $n(x,y)$ represents the surface normal, and $R(x,y)$ is the irradiance map for lighting direction $l$.

Diffuse shading can be expressed as an integral of the radiance over all the angular directions. For discrete samples of the angular directions, the diffuse shading map will be a weighted sum of the irradiance maps for the sampled directions, with the light intensity as the weight.
As a result, with an accurate shading map and irradiance maps, we can directly solve for the lighting via optimization \cite{Sato2003IFS,Barron2015}. However, optimization based solutions suffer from inaccurate shading and geometry and slow computation. 
Our translation also follows such a physically-based design. The neural network takes the same input as the numerical optimization, namely the diffuse shading map and the
irradiance maps of sparsely sampled lighting directions. Then the network outputs the intensity values of those sampled lighting directions. In practice, we sample the pixel centers of a $8\times 8\times 6$ cube-map as the lighting directions to compute a $32 \times 32$ resolution auxiliary irradiance map. The diffuse shading map and each irradiance map is separately passed
through one layer of convolution, then their feature maps are concatenated together. The concatenated features are sent through a series of convolution and pooling layers to produce a set of feature maps corresponding to $8 \times 8 \times 6$ angular directions. With the feature maps now defined in the angular domain, we rearrange them into a latitude and longitude representation and perform a series of angular domain convolutions to output an estimated environment map of $256 \times 128$ resolution.

\paragraph{Specular shading translation}
Unlike the lowpass filtering effects of diffuse shading, specular reflections are generally of much higher frequency and cannot be adequately represented by the sparse low-resolution sampling used for diffuse shading translation, making the irradiance based solution inefficient.
Since specular reflection has a strong response along the mirror direction,
we thus map the decomposed specular shading to the angular domain according to the mirror direction, computed from the viewing direction and surface normal as $o = (2n - v)$. Each pixel in the specular shading map is translated individually, and the average value is computed among all the pixels that are mapped to the same angular map location, represented in terms of latitude and longitude.

In practice, depending on the scene geometry, some angular directions may not have any corresponding pixels in the specular shading map. To take this into account, we maintain an additional mask map that indicates whether an angular direction has a valid intensity value or not. After mapping to the $256 \times 128$ latitude and longitude map, the maps are processed by four convolution layers to produce angular feature maps.

\paragraph{Angular fusion network}
\label{sec:fusion}
After the diffuse and specular translation network, both the diffuse and specular feature maps are defined in the angular domain. Those feature maps will be concatenated together for the fusion network to determine the final lighting estimates.

The angular fusion network is a standard U-net structure, with a series of convolutions and downsampling for encoding, followed by convolution and upsampling layers for decoding. We also include skip links to map the feature maps from the encoder to the decoder, to better preserve angular details.
Instead of having the estimated lighting from the diffuse spatial-angular translation network as input, we use the feature map (just before the output layer) of the translation network as the diffuse input to the fusion network. The feature map can preserve more information processed by the translation network (e.g. confidence of the current estimation / transformation), which may help with fusion but is lost in the lighting output.

\paragraph{Recurrent convolution layers}
\label{sec:recurrent}
Due to the ill-conditioned nature of lighting estimation, there may exist ambiguity in the solution for a video frame. Moreover, this ambiguity may differ from frame to frame because of the different information that the frames contain (e.g., specular objects reflect light from different directions depending on the viewing angle, and cast shadows might be occluded by the objects in certain views). To help resolve these ambiguities and obtain a more coherent solution, we make use of temporal information from multiple frames in determining light predictions.
This is done by introducing recurrent convolution layers into the diffuse shading translation network as well as the angular fusion network.

\paragraph{Depth estimation for RGB only input}
Our system is built with RGBD input in mind. However, scene depth could instead be estimated by another neural network, which would allow our system to run on RGB images without measured depth. We trained a depth estimation network adapted from \cite{Laina2016} but with the viewpoint information as an additional input, as this may facilitate estimation over the large range of pitch directions encountered in AR scenarios. The estimated depth is then included as part of our input.
Figure \ref{fig:paul_comp} presents real measured examples with light estimated using the predicted depth.

%% file: training.tex
\section{Training}
\label{sec:training}

\subsection{Supervision and training losses}
We train all the networks using synthetically generated data to provide supervision for each network individually. For the reflectance decomposition network, ground truth maps are used to supervise its training. Following previous networks for intrinsic image decomposition \cite{Li:2018:CGI}, we employ the L2 loss of the maps as well as the L2 loss of the gradients of the albedo and specular maps. The fusion network is trained with the ground truth full-resolution environment map, using the Huber loss.

\paragraph{Recurrent convolution training}
We train the recurrent convolution layers with sequential data and expand the recurrent layers. In practice, we expand the recurrent layers to accommodate $10$ consecutive frames during training.
When training the recurrent layers, in addition to the Huber loss that minimizes the difference between the estimated lighting and the ground truth for each frame, we also include a temporal smoothness loss to promote smoothness between consecutively estimated lights $L_i, L_{i+1}$, defined as
\begin{equation}
    \mathcal{L}_{t}(L_i, L_{i+1}) = || L_i - \mathcal W_{i+1 \to i}(L_{i+1}) ||^2
\end{equation}
where $W_{i+1 \to i}$ is the ground truth optical flow from the lighting of the $(i+1)$-th frame to the $i$-th frame.

\subsection{Training data preparation}

The training data is composed of many elements such as reflectance maps and angular environment lighting which are difficult to collect in large volumes from real scenes. As a result, we choose to train our neural networks with synthetically generated data. For better network generality to real-world inputs, this synthetic data is prepared to better reflect the properties of real scenes.
The environment lights consist of 500 HDR environment maps collected from the Internet, 500 HDR environment map captured by a 360 camera and 14K randomly generated multiple area light sources. 
We collected over 600 artist modeled objects with realistic texture details and material variations and organize those objects into local scenes that include about 1 - 6 objects. The scene is then lit by randomly selected environment lights. The viewpoint is uniformly sampled over the upper hemisphere to reflect most AR scenarios. Please refer to the supplementary materials for more details about training data preparation.

\subsection{Implementation}
We implement the neural networks in Tensorflow \cite{tensorflow2015-whitepaper}.
We train our system using the Adam \cite{Kingma:2015:AMS} optimizer, with a $10^{-5}$ learning rate and the default settings for other hyper-parameters. The decomposition is separately trained using $2 \times 10^6$ iterations; the diffuse-translation network and the fusion network are first trained with $8 \times 10^{5}$ iterations without recurrent layers, and then finetuned over $10^5$ iterations for the recurrent layer training. Finally, the whole system is finetuned end-to-end with $10^5$ iterations. Training the decomposition network takes about 1 week on a single NVidia Titan X GPU. The diffuse-translation and fusion network training takes about $10$ hours, and the final end-to-end finetuning takes about $20$ hours.

%% file: results.tex
\input{fig_light_env}

\section{Results}
\label{sec:results}

\subsection{Validations}
\paragraph{Single-image inputs}
Beside video frames, our network does support single-image input by feeding ten copies of the single static image into the network to get the lighting estimation result. This allows us to perform validation and comparsion to existing works on single-image inputs. 

\paragraph{Error metric}
To measure the accuracy of the estimated light, we directly measure the RMSE compared to the reference environment map. In addition, since the goal of our lighting estimation is rendering virtual objects for AR applications, we also measure the accuracy of our system using the rendering error of a fixed set of objects with the estimated light. All the objects in this set are shown as virtual inserts in our real measured results. 

\input{fig_light_size}
\paragraph{Dataset}
To evaluate the performance of our method on real inputs, we captured 180 real input images with the ground truth environment map, providing both numerical and visual results.
To systematically analyze how our method works on inputs with different light, layout and materials, we also designed a comprehensive synthetic test set with full control of each individual factor.
Specifically, we collected eight artist-designed scenes to ensure plausible layouts, various object shapes, and a range of objects. Ten random viewpoints are generated for each scene, under selected environment maps with random rotations. 

\paragraph{Real measured environment maps}
For testing the performance of our method under different environment lights, we collected 20 indoor environment maps which are not used for neural network training. These environment maps can be classified into several categories that represent different kinds of lighting conditions. Specifically, we have five environment maps that contain a single dominant light source, five with multiple dominant lights, five with very large area light sources, and five environment maps with near ambient lighting.

\input{tab_analysis}
\input{fig_material_plot}
\input{fig_sh_gi}

Table~\ref{tab:analysis} lists the average error for each category of environment light.
Intuitively, lights with high frequency details, such as a dominant light, are more challenging than low frequency cases, like near-ambient lighting.
Visual results for one typical example from each category are displayed in Figure~\ref{fig:env}, which shows that our method can correctly estimate environment lighting and produce consistent re-rendering results across all the types of environment lighting.

\paragraph{Synthetic environment maps}
We also synthetically generate a set of lighting conditions that provide more continuous coverage over area light source size and number of light sources. For each set of lighting conditions, we render each scene and each view under those lighting conditions with ten random rotations. We then plot the average error of our system on the different lighting conditions in the set.

Figure~\ref{fig:material_plot} plots the lighting estimates and re-rendering error for each dataset.
Inputs with smaller area light sources are more challenging for lighting estimation, since small changes of light position leads to large error in both the angular lighting domain as well as the re-rendering results. Figure~\ref{fig:light_size_num} (a-d) exhibits a selected scene under various lighting conditions.
For a near-diffuse scene lit by multiple area light sources, even the shadows of objects do not provide sufficient information to fully recover the rich details of each individual light source. However, the re-rendering results with our estimated lighting matches well with that rendered with ground truth light, making them plausible for most AR applications.

\paragraph{Object materials}
To analyze the how surface material affects our estimation results, we prepared a set of objects with the same shape but varying materials.
We use a homogeneous specular BRDF with varying roughness and varying diffuse-specular ratios.
The results, shown in Figure~\ref{fig:light_size_num} (e-h), indicate that lighting estimation is stable to a wide range of object materials, from highly specular to purely diffuse. Since our method uses both shading and shadow cues in the scene for lighting estimation, the lighting estimation error is not sensitive to the material of the object, with only a slightly larger error for a purely diffuse scene.

\paragraph{Layouts}
We test the performance of our method on scenes with different layouts by classifying our scenes based on complexity. The numerical results for the two layout categories are listed in Table~\ref{tab:analysis}. As expected, complex scene layouts lead to more difficult lighting estimation. However, as shown in Figure~\ref{fig:light_size_num} (i,j), our method produces high quality results in both cases.

\paragraph{Spatially-varying illumination}
Our object-based lighting estimation can be easily extended to support spatially-varying illumination effects, such as near-field illumination, by taking different local regions of the input image. Please find example results in the supplementary material.  

\input{fig_ourrealcomp}

\input{fig_paulcomp}
\input{fig_redwood}

\subsection{Comparisons}

Here, we compare our method to existing lighting estimation methods. For systematic evaluation with ground truth data, we compared with scene-based methods \cite{Gardner2017,Gardner_2019_ICCV} as well as an inverse rendering method \cite{Sato2003IFS}, on both synthetic and real measured test sets.
Our method takes advantage of physical knowledge to facilitate inference from various cues in the input data, and outperforms all the existing methods, as shown in Table~\ref{tab:analysis}.
In Figure \ref{fig:ourrealcomp}, we show one example of a real captured example. Note that our method estimates an area light source with the right size at the correct location, while other methods either over-estimate the ambient light \cite{Gardner2017} or result in an incorrect light position \cite{Gardner_2019_ICCV,Sato2003IFS}.

We also compare our method on the redwood RGBD dataset \cite{Choi2016}, as illustrated in Figure \ref{fig:redwoodreal}. Although without a ground truth reference, virtual objects rendered with our estimated lights produce consistent specular highlights and shadows, while existing methods fail to generate consistent renderings.

Scene based methods usually need input with a wider field of view, thus we compare our method with \cite{LeGendre:2019:DLI,Gardner2017} on their ideal input. Their methods use the full image as input, while our method uses only a small local crop around the target object. The depth of the crop is estimated for input to our system. Figure \ref{fig:paul_comp} illustrates one example result. Note that the rendering result with our estimated light matches well to the reference (the highlight should be at the left side of the bunny).
More comparisons can be found in the supplementary material.

To compare with methods based on low-order spherical harmonics (SH) \cite{Garon_2019_CVPR}, we fit the ground-truth light with 3- and 5-order SH and compare the light and rendering results with our method. As shown in Figure \ref{fig:shgi}, a 5-order SH (as used by \cite{Garon_2019_CVPR}) is incapable of representing small area light sources and generates blurred shadow; our method estimates a full environment map and produces consistent shadow in the rendering.

\subsection{Ablation studies}
\label{sec:ablations}
We conduct ablation studies to justify the components of our system, including the separation into diffuse and specular reflections, and to verify the effectiveness of the translation networks.

We first test our system without the diffuse or specular translation network, but with the remaining networks unchanged. Empirical comparisons are shown in Figure~\ref{fig:ablation_net}. Without the specular translation network, the system fails to estimate the high-frequency lighting details and produces inconsistent specular highlights in the re-rendering results. Without the diffuse translation network, the system found difficulties estimating light from a diffuse scene.

We then remove the decomposition network and feed the RGBD input directly to the spatial-angular translation networks, followed by the fusion network.
It is also possible to train a neural network to regress the angular environment lighting directly from the RGBD input. Such a neural network structure shares a design similar to \cite{GeorgoulisPAMI2017,LeGendre:2019:DLI} but is trained on our dataset. As shown in Table~\ref{tab:analysis} (numerical) and Figure~\ref{fig:ablation_net} (visual), compared to those alternative solutions, our method produces the best quality results by combining both diffuse and specular information with our physics-aware network design.
With only RGB input, our method can also predict environment maps with estimated depth, and outperforms existing methods. Larger estimation error is found due to inaccuracy in the depth input, which could be improved by training our network together with the depth estimation network.

Please refer to our supplementary video to see how the recurrent convolution layers increase the temporal coherence of the estimated illumination.

\input{fig_ablation_net}
Finally, we captured sequences of indoor scenes with dynamic lighting effects, and continuously changing viewpoints, demonstrating the advantages of having real-time lighting estimation for AR applications. Please refer to the supplementary material for the implementation details and real video sequence results. For all the real video results, we crop the central $384 \times 384$ region as the input of our method.

\subsection{Performance}
We test the runtime performance of our method on a workstation with an Intel 7920x CPU and a single NVidia RTX 2080 Ti GPU. For the full system, the processing time per frame is within 26 ms, of which 14 ms is needed for preparing the input and 12 ms is used for neural network inference. We regard adopting to mobile GPUs as future work.

%% file: fig_light_env.tex

\begin{figure}[b]
   \centering
\bgroup
\scriptsize
\renewcommand{\tabcolsep}{0pt}
\renewcommand{\arraystretch}{0.2}
\begin{tabular}{m{1em} m{\fourColumnFigWidth}m{\fourColumnFigWidth} c m{\fourColumnFigWidth}m{\fourColumnFigWidth} c m{\fourColumnFigWidth}m{\fourColumnFigWidth}  c m{\fourColumnFigWidth}m{\fourColumnFigWidth}}
& \multicolumn{3}{c}{(a) One dominant} & \multicolumn{3}{c}{(b) Multiple lights} & \multicolumn{3}{c}{(c) Large area light} & \multicolumn{2}{c}{(d) Near ambient}\\
 \rotatebox{90}{Render}&
\includegraphics[width=\fourColumnFigWidth]{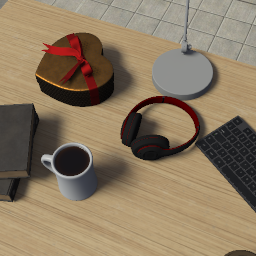}&
\includegraphics[width=\fourColumnFigWidth]{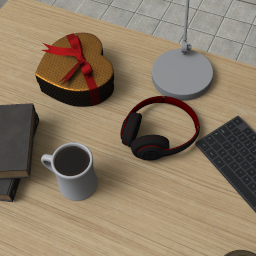}& \hspace{2pt} &
\includegraphics[width=\fourColumnFigWidth]{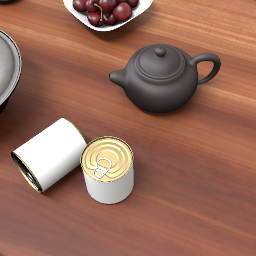}&
\includegraphics[width=\fourColumnFigWidth]{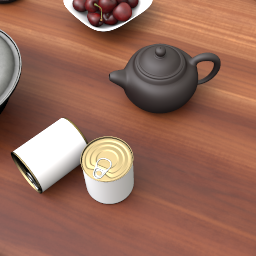}& \hspace{2pt} &
\includegraphics[width=\fourColumnFigWidth]{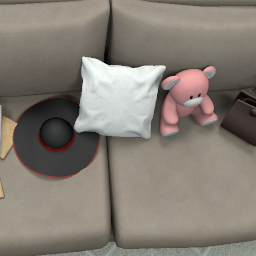}&
\includegraphics[width=\fourColumnFigWidth]{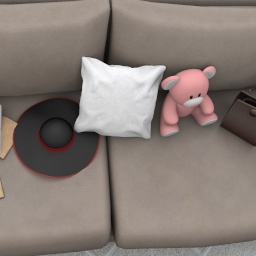}& \hspace{2pt} &
\includegraphics[width=\fourColumnFigWidth]{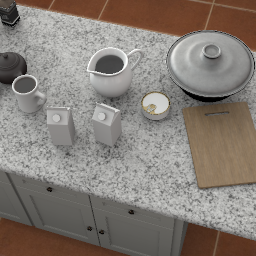}&
\includegraphics[width=\fourColumnFigWidth]{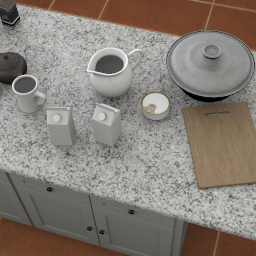}\\
 \rotatebox{90}{Light}&
 \includegraphics[width=\fourColumnFigWidth]{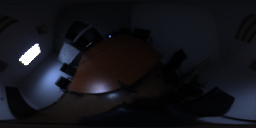}&
\includegraphics[width=\fourColumnFigWidth]{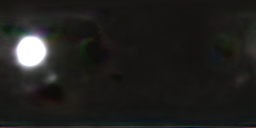}& \hspace{2pt} &
\includegraphics[width=\fourColumnFigWidth]{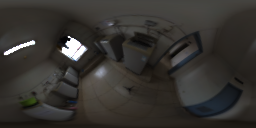}&
\includegraphics[width=\fourColumnFigWidth]{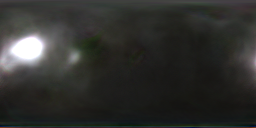}& \hspace{2pt} &
\includegraphics[width=\fourColumnFigWidth]{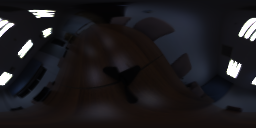}&
\includegraphics[width=\fourColumnFigWidth]{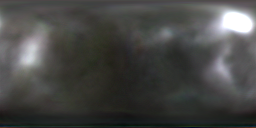}& \hspace{2pt} &
\includegraphics[width=\fourColumnFigWidth]{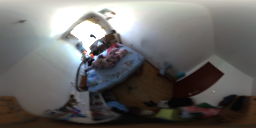}&
\includegraphics[width=\fourColumnFigWidth]{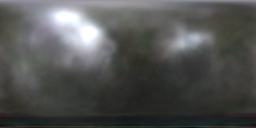}
\end{tabular}
\egroup
    \vspace*{-0.4cm}
    \caption{We test our system with different classes of environment maps. For each class, the left column shows the ground truth, and the right column shows our results.}
    \label{fig:env}
\end{figure}

%% file: fig_light_size.tex
\newcommand{\newlayoutWidth}{1.2cm}

\begin{figure*}
   \centering
\bgroup
\tiny
\renewcommand{\tabcolsep}{0pt}
\renewcommand{\arraystretch}{0.2}
\begin{tabular}{m{\newlayoutWidth}m{\newlayoutWidth} c m{\newlayoutWidth}m{\newlayoutWidth} c m{\newlayoutWidth}m{\newlayoutWidth}  c m{\newlayoutWidth}m{\newlayoutWidth} c m{\newlayoutWidth}m{\newlayoutWidth}}
\multicolumn{3}{c}{(a) Small area light} & \multicolumn{3}{c}{(b) Middle area light} & \multicolumn{3}{c}{(c) Large area light} & \multicolumn{3}{c}{(d) Multiple area lights} & \multicolumn{2}{c}{(i) Simple layout}\\
\includegraphics[width=\newlayoutWidth]{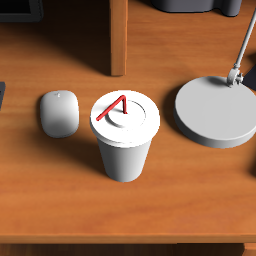}&
\includegraphics[width=\newlayoutWidth]{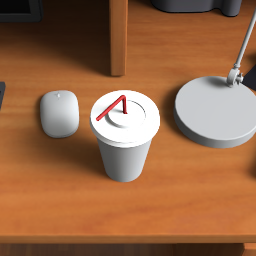}& \hspace{1pt} &
\includegraphics[width=\newlayoutWidth]{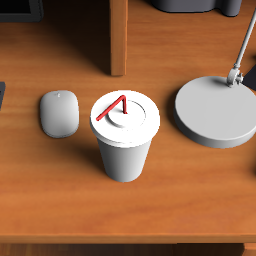}&
\includegraphics[width=\newlayoutWidth]{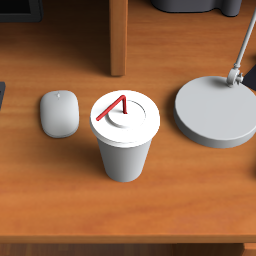}& \hspace{1pt} &
\includegraphics[width=\newlayoutWidth]{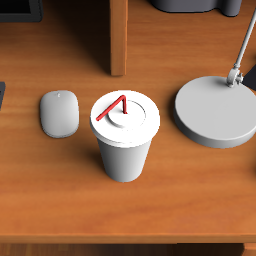}&
\includegraphics[width=\newlayoutWidth]{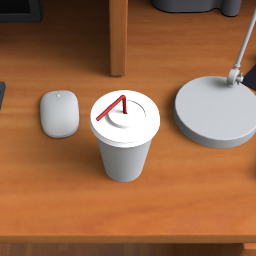}& \hspace{1pt} &
\includegraphics[width=\newlayoutWidth]{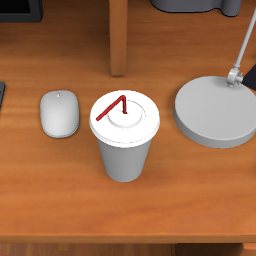}&
\includegraphics[width=\newlayoutWidth]{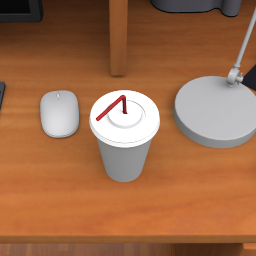}& \hspace{3pt} &
\includegraphics[width=\newlayoutWidth]{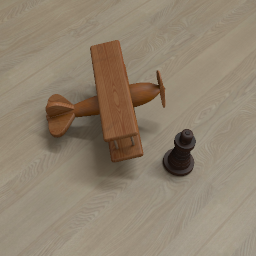}&
\includegraphics[width=\newlayoutWidth]{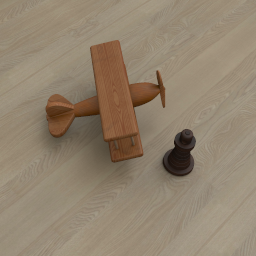}\\
 \includegraphics[width=\newlayoutWidth]{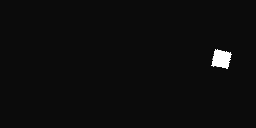}&
\includegraphics[width=\newlayoutWidth]{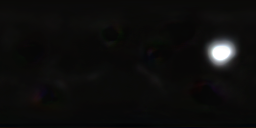}& \hspace{1pt} &
\includegraphics[width=\newlayoutWidth]{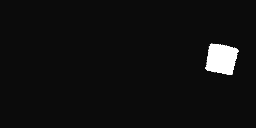}&
\includegraphics[width=\newlayoutWidth]{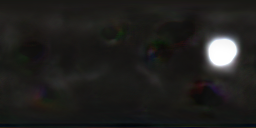}& \hspace{1pt} &
\includegraphics[width=\newlayoutWidth]{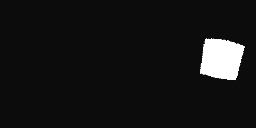}&
\includegraphics[width=\newlayoutWidth]{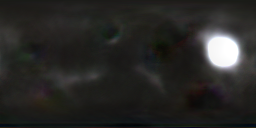}& \hspace{1pt} &
\includegraphics[width=\newlayoutWidth]{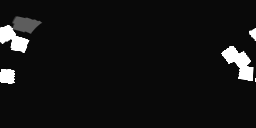}&
\includegraphics[width=\newlayoutWidth]{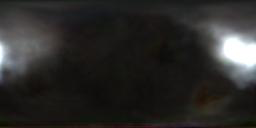}& \hspace{3pt} &
\includegraphics[width=\newlayoutWidth]{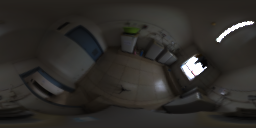}&
\includegraphics[width=\newlayoutWidth]{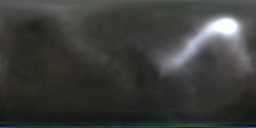} \\

\multicolumn{3}{c}{(e) $\rho_s 0.5, \rho_d 0.5, \sigma 0.05$} & \multicolumn{3}{c}{(f) $\rho_s 0.5, \rho_d 0.5, \sigma 0.1$} & \multicolumn{3}{c}{(g) $\rho_s 0.1, \rho_d 0.9, \sigma 0.1$} & \multicolumn{3}{c}{(h) $\rho_s 0.0, \rho_d 1.0, \sigma 0.1$} & \multicolumn{2}{c}{(j) Complex layout}\\

\includegraphics[width=\newlayoutWidth]{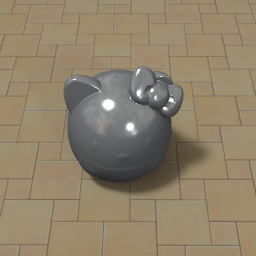}&
\includegraphics[width=\newlayoutWidth]{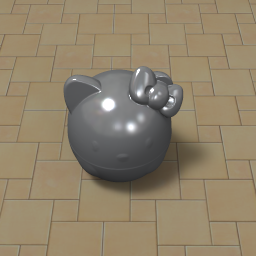}& \hspace{1pt} &
\includegraphics[width=\newlayoutWidth]{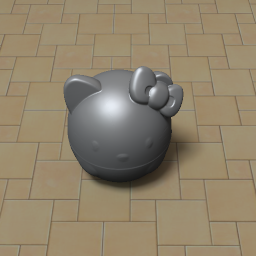}&
\includegraphics[width=\newlayoutWidth]{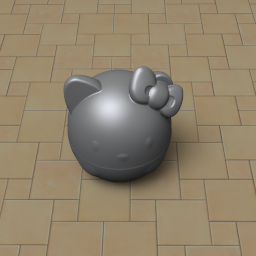}& \hspace{1pt} &
\includegraphics[width=\newlayoutWidth]{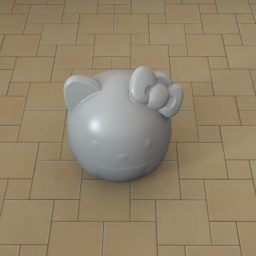}&
\includegraphics[width=\newlayoutWidth]{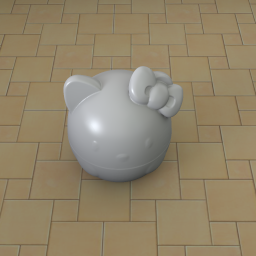}& \hspace{1pt} &
\includegraphics[width=\newlayoutWidth]{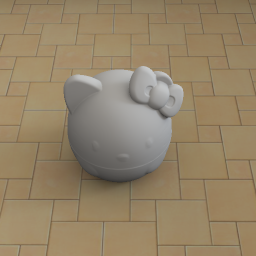}&
\includegraphics[width=\newlayoutWidth]{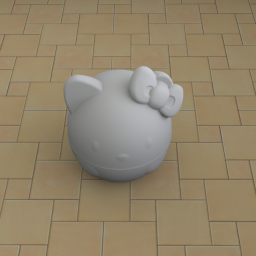}& \hspace{3pt} &
\includegraphics[width=\newlayoutWidth]{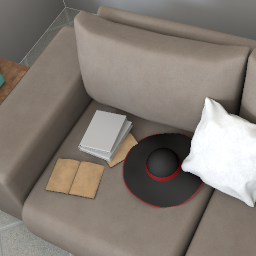}&
\includegraphics[width=\newlayoutWidth]{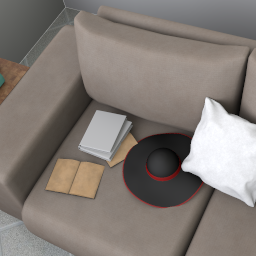}\\
\includegraphics[width=\newlayoutWidth]{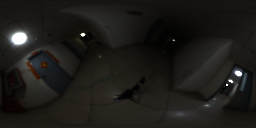}&
\includegraphics[width=\newlayoutWidth]{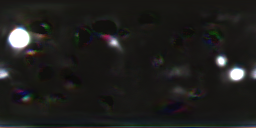}& \hspace{1pt} &
\includegraphics[width=\newlayoutWidth]{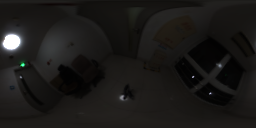}&
\includegraphics[width=\newlayoutWidth]{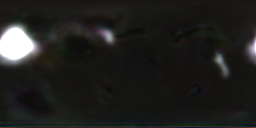}& \hspace{1pt} &
\includegraphics[width=\newlayoutWidth]{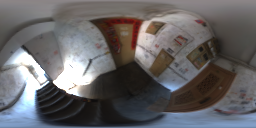}&
\includegraphics[width=\newlayoutWidth]{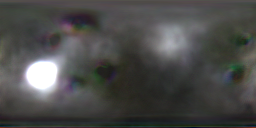}& \hspace{1pt} &
\includegraphics[width=\newlayoutWidth]{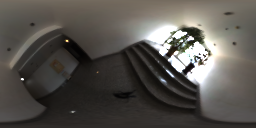}&
\includegraphics[width=\newlayoutWidth]{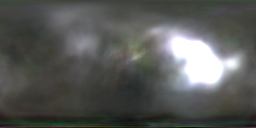}& \hspace{3pt} &
\includegraphics[width=\newlayoutWidth]{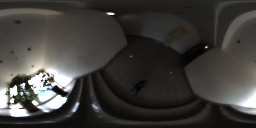}&
\includegraphics[width=\newlayoutWidth]{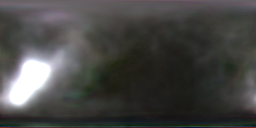} \\

\end{tabular}
\egroup
    \caption{(a-d) Light estimation results for the same scene lit by different-sized area light sources and multiple light sources.
    (e-h) Light estimation results for the same shape with varying material properties.
    (i,j) Light estimation results for scenes with simple or complex layouts.
    In each group, the left column shows the ground truth, and the right column shows our results.}
    \label{fig:light_size_num} 
\end{figure*}

%% file: tab_analysis.tex

\begin{table}

\bgroup
\scriptsize
\begin{minipage}[t]{0.35\textwidth}
\centering

\begin{tabular}{ l | c c }
    & Render & Light \\
\hline
\multicolumn{3}{c}{Environment light}\\
\hline
    One dominant & 0.057        &2.163 \\
Multiple lights  & 0.057        &1.653 \\
Large area light & 0.045        &1.222 \\
Near ambient     & 0.038        &0.994 \\
\hline
\multicolumn{3}{c}{Layout robustness}\\
\hline
Simple layout    & 0.046        &0.915 \\
Complex layout   & 0.049        &1.508 \\
\hline
\multicolumn{3}{c}{Real captured test set}\\
\hline
Real captured & 0.052 & 0.742 \\
\hline
  \end{tabular}
\end{minipage}
\begin{minipage}[t]{0.2\textwidth}
\centering
\begin{tabular}{ l | c c | c c  }
    & \multicolumn{2}{c|}{Synthetic} & \multicolumn{2}{c}{Real} \\
\hline
\multicolumn{1}{c|}{Comparasion} & Render & Light & Render & Light\\
\hline
Gardner et al. 17\cite{Gardner2017} & 0.095 & 3.056 & 0.088 & 3.462\\
Gardner et al. 19\cite{Gardner_2019_ICCV} & 0.100 & 1.614 & 0.093 & 1.303\\
Sato\cite{Sato2003IFS} + our decomposition & 0.073 & 1.851 & 0.081 & 1.257\\
Sato\cite{Sato2003IFS} + ground truth shading & 0.064 & 1.944 & - & - \\
\hline
\multicolumn{1}{c|}{Ablation}\\
\hline
Diffuse only         & 0.050        & 1.447 & 0.055 & 0.768\\
Specualr only        &0.056     & 1.440  & 0.061 & 0.756\\
Without decomposition& 0.052        & 1.434 & 0.059 & 0.794\\
Direct regression    & 0.049        & 1.432 & 0.057 & 0.759 \\
Without recurrent layer  & 0.048 &  1.429 & 0.056 & 0.760 \\
\hline
Our results & \textbf{0.046}    & \textbf{1.419} & \textbf{0.052} & \textbf{0.742} \\
Our results (estimated depth)    & 0.053    & 1.437 & 0.064 & 0.773\\
    \hline
  \end{tabular}

\end{minipage}
\egroup
\caption{Average RMSE of estimated lights and re-rendered images.   }
\label{tab:analysis}
\end{table}

%% file: fig_material_plot.tex

\begin{figure}
   \centering\includegraphics[width=\textwidth]{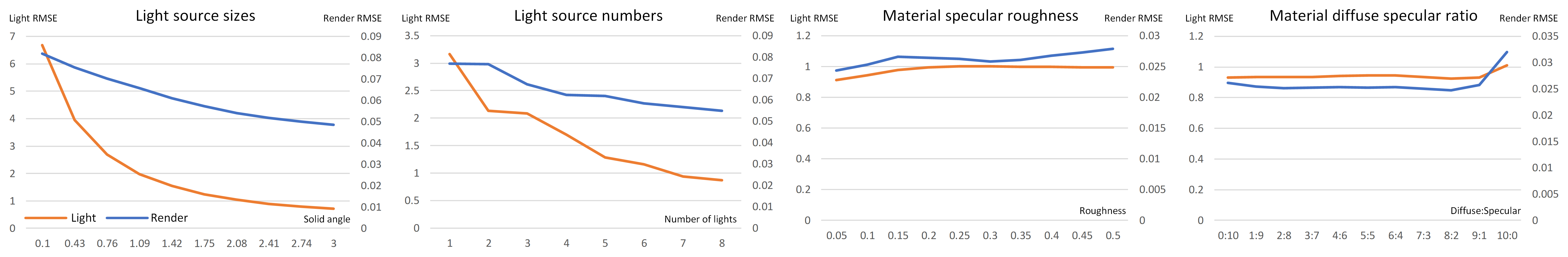}
    \caption{Lighting estimation and re-rendering error with respect to area light source sizes, number of light sources and different surface materials. }
    \label{fig:material_plot}
\end{figure}

%% file: fig_sh_gi.tex

\begin{figure}
   \centering
\bgroup
\tiny
\renewcommand{\tabcolsep}{0pt}
\renewcommand{\arraystretch}{0.2}
\begin{tabular}{m{0.8em} m{\fourColumnFigWidth}m{\fourColumnFigWidth} m{\fourColumnFigWidth}m{\fourColumnFigWidth} c m{\fourColumnFigWidth}m{\fourColumnFigWidth}  m{\fourColumnFigWidth}m{\fourColumnFigWidth}}
& \multicolumn{1}{c}{(a)Input } & \multicolumn{1}{c}{(b)Ours } & \multicolumn{1}{c}{(c)3-order SH} & \multicolumn{1}{c}{(d)5-order SH} & & \multicolumn{1}{c}{(e)Input} & \multicolumn{1}{c}{(f)Ours} & \multicolumn{1}{c}{(g)3-order SH} & \multicolumn{1}{c}{(h)5-order SH}\\
 \rotatebox{90}{Render}&
\includegraphics[width=\fourColumnFigWidth]{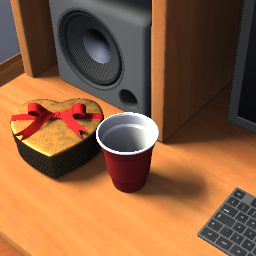}&
\includegraphics[width=\fourColumnFigWidth]{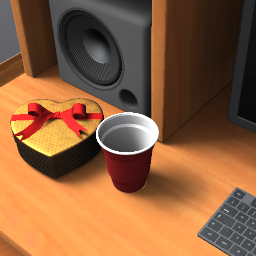}&
\includegraphics[width=\fourColumnFigWidth]{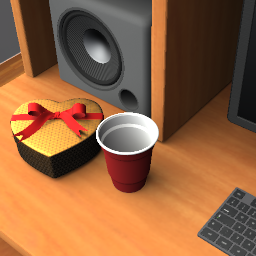}&
\includegraphics[width=\fourColumnFigWidth]{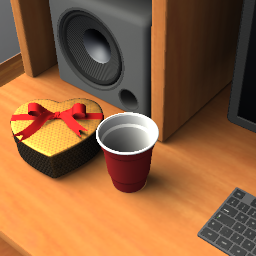}& \hspace{1pt} &
\includegraphics[width=\fourColumnFigWidth]{fig/env/01/render_gt.png}&
\includegraphics[width=\fourColumnFigWidth]{fig/env/01/render_pred.png}&
\includegraphics[width=\fourColumnFigWidth]{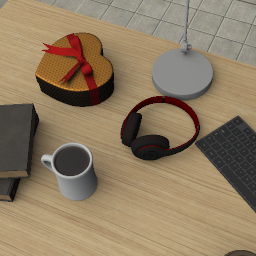}&
\includegraphics[width=\fourColumnFigWidth]{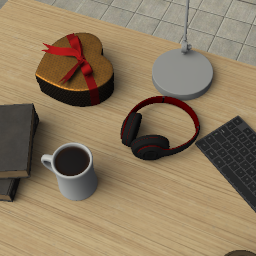}\\
 \rotatebox{90}{Light}&
\includegraphics[width=\fourColumnFigWidth]{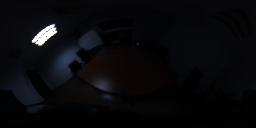}&
\includegraphics[width=\fourColumnFigWidth]{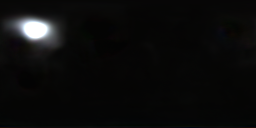}&
\includegraphics[width=\fourColumnFigWidth]{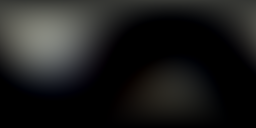}&
\includegraphics[width=\fourColumnFigWidth]{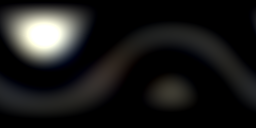}& \hspace{1pt} &
\includegraphics[width=\fourColumnFigWidth]{fig/env/01/light_gt.png}&
\includegraphics[width=\fourColumnFigWidth]{fig/env/01/light_pred.png}&
\includegraphics[width=\fourColumnFigWidth]{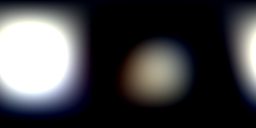}&
\includegraphics[width=\fourColumnFigWidth]{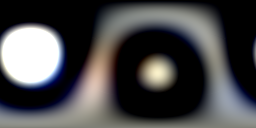}
\end{tabular}
\egroup
    \caption{Comparison to SH-based representation. Our method recovers an all-frequency environment map and produces sharp shadows similar to the ground truth. Fitting the ground truth light with 3- or 5-order SH still cannot reproduce such effects. }
    \label{fig:shgi} 
\end{figure}

%% file: fig_ourrealcomp.tex

\newcommand{\ourrealcompSize}{1.9cm}

\begin{figure}
   \centering
\bgroup
\tiny
\renewcommand{\tabcolsep}{0pt}
\renewcommand{\arraystretch}{0.2}
\begin{tabular}{m{\ourrealcompSize}m{\ourrealcompSize}m{\ourrealcompSize}m{\ourrealcompSize}m{\ourrealcompSize}m{\ourrealcompSize}}
\multicolumn{1}{c}{(a) Input} & \multicolumn{1}{c}{(b) Reference} & \multicolumn{1}{c}{(c) Ours} & \multicolumn{1}{c}{(d) \protect{\cite{Gardner2017}} } & \multicolumn{1}{c}{(e) \protect{\cite{Gardner_2019_ICCV}}} & \multicolumn{1}{c}{(f)\protect{\cite{Sato2003IFS}}+our decom. } \\

\includegraphics[width=\ourrealcompSize]{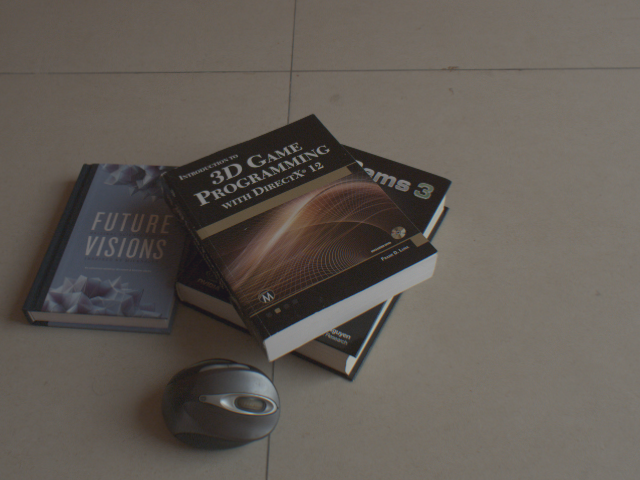}&
\includegraphics[width=\ourrealcompSize]{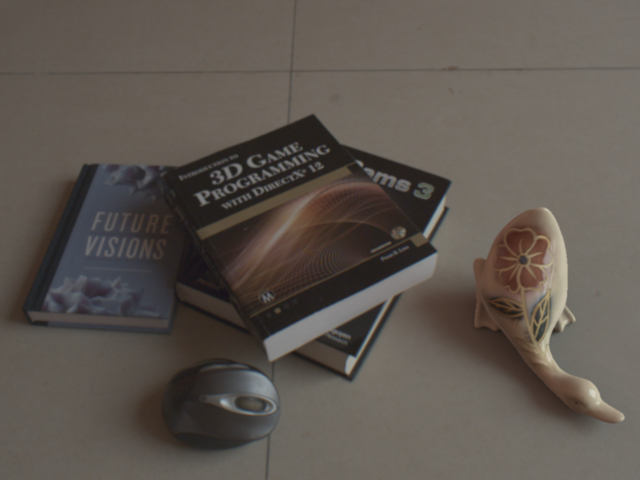}& 
\includegraphics[width=\ourrealcompSize]{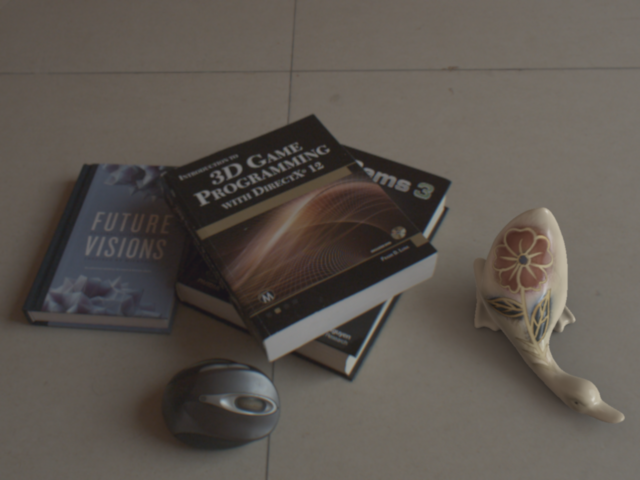}&
\includegraphics[width=\ourrealcompSize]{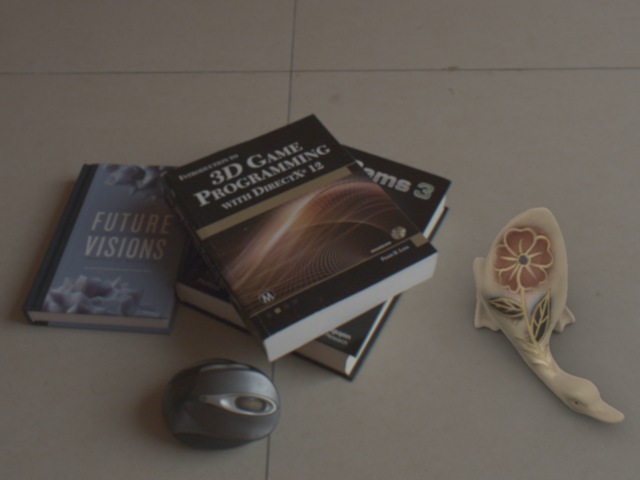}&
\includegraphics[width=\ourrealcompSize]{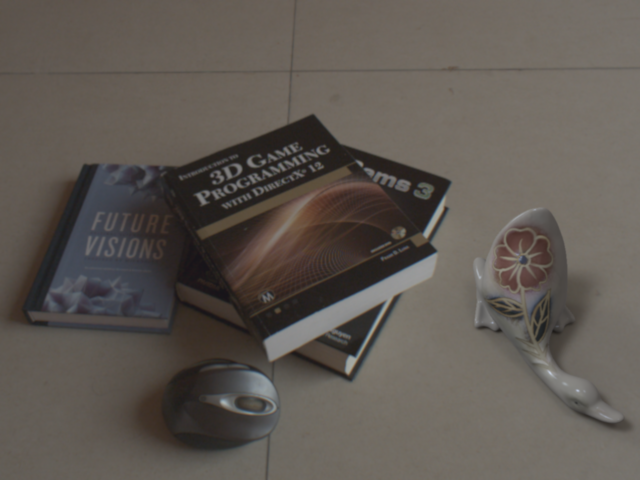}&
\includegraphics[width=\ourrealcompSize]{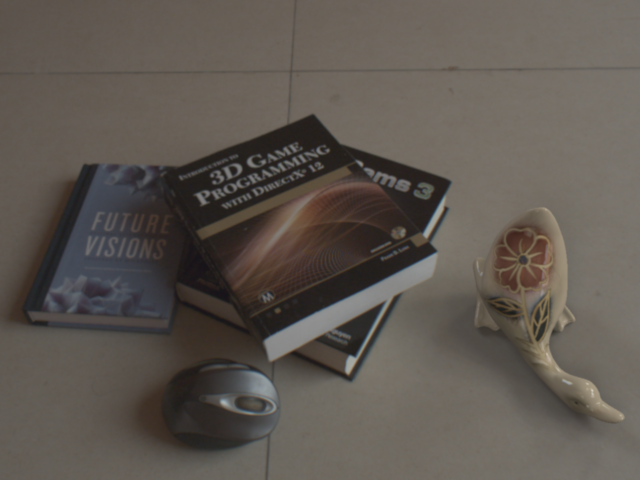}\\
 &
\includegraphics[width=\ourrealcompSize]{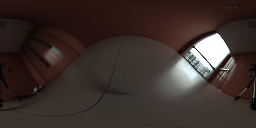}& 
\includegraphics[width=\ourrealcompSize]{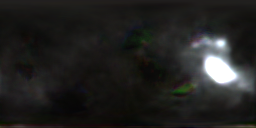}&
\includegraphics[width=\ourrealcompSize]{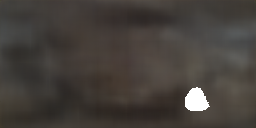}&
\includegraphics[width=\ourrealcompSize]{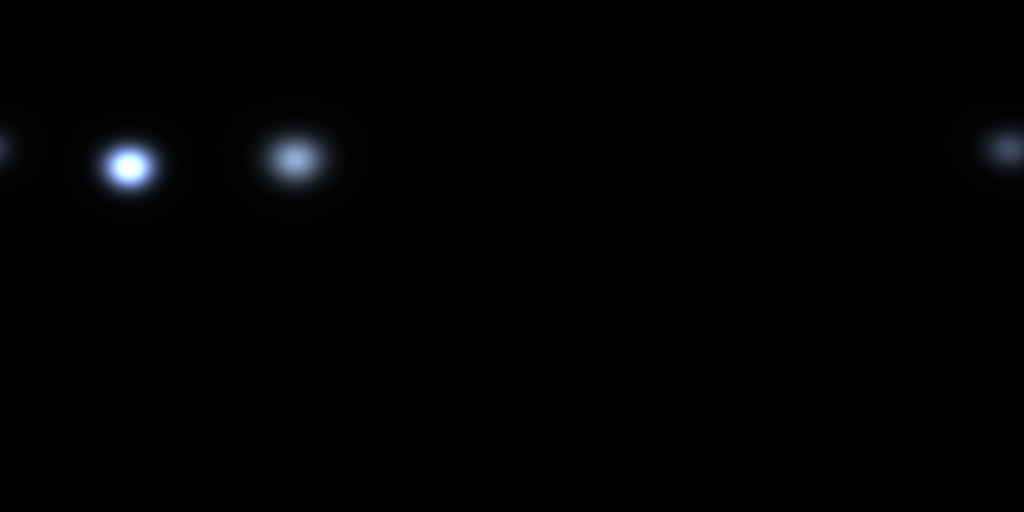}&
\includegraphics[width=\ourrealcompSize]{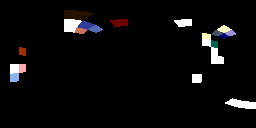}\\

\includegraphics[width=\ourrealcompSize]{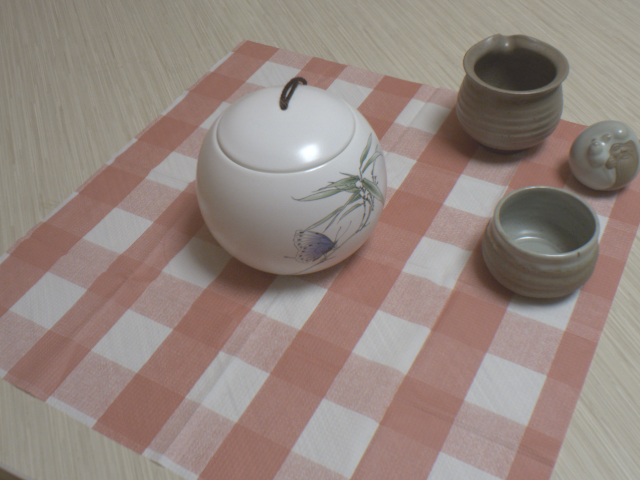}&
\includegraphics[width=\ourrealcompSize]{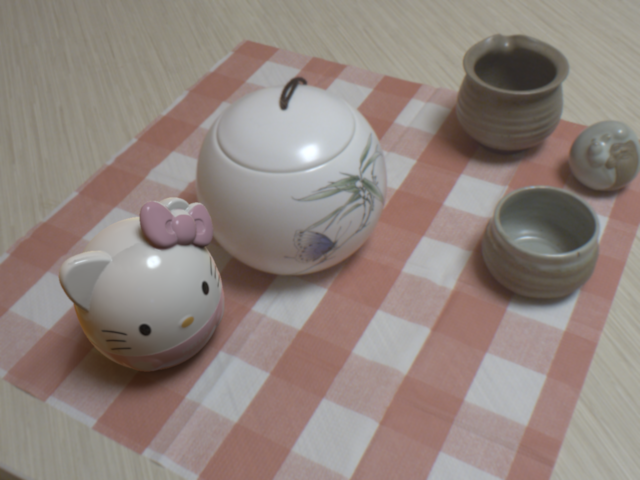}& 
\includegraphics[width=\ourrealcompSize]{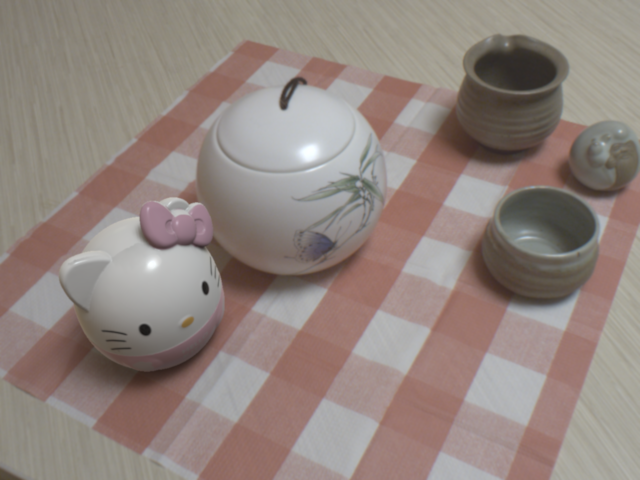}&
\includegraphics[width=\ourrealcompSize]{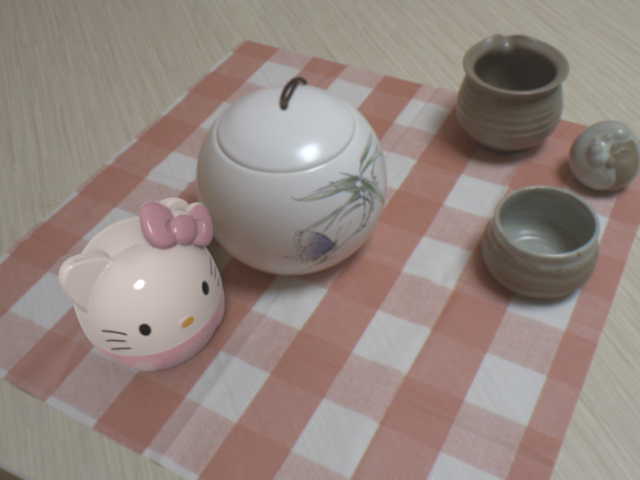}&
\includegraphics[width=\ourrealcompSize]{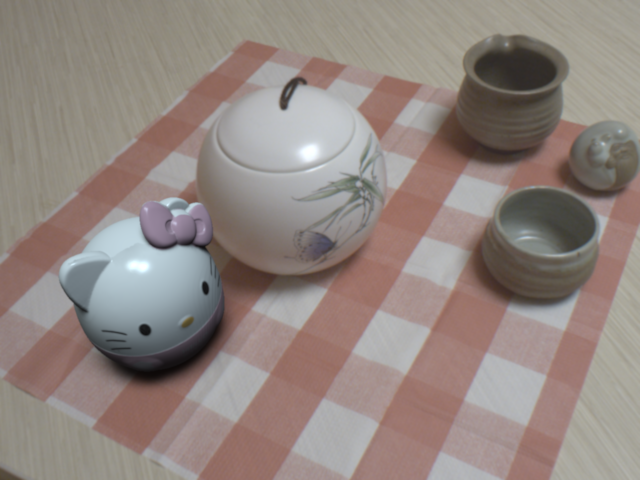}&
\includegraphics[width=\ourrealcompSize]{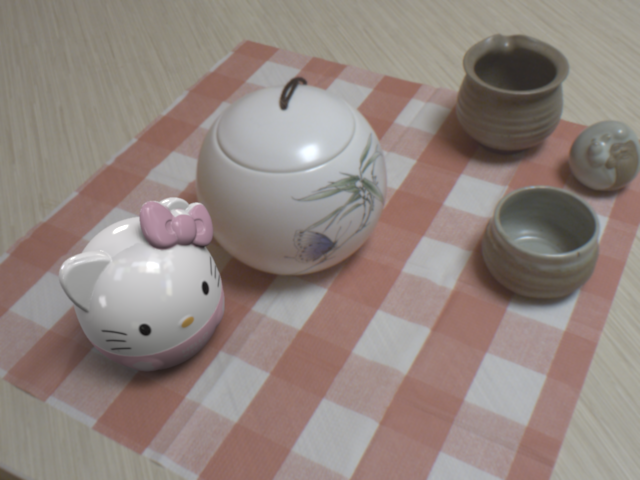}\\
 &
\includegraphics[width=\ourrealcompSize]{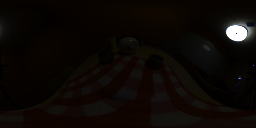}& 
\includegraphics[width=\ourrealcompSize]{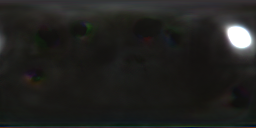}&
\includegraphics[width=\ourrealcompSize]{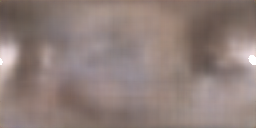}&
\includegraphics[width=\ourrealcompSize]{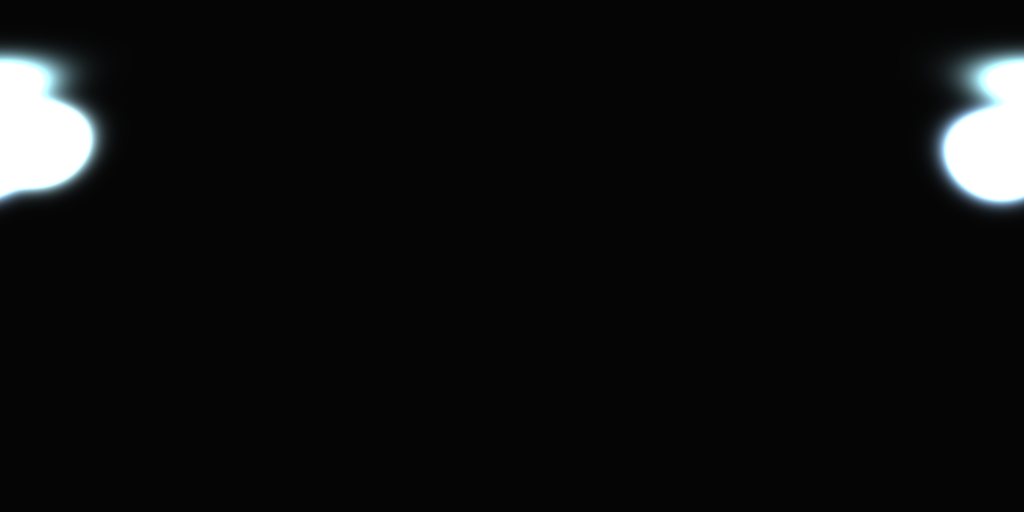}&
\includegraphics[width=\ourrealcompSize]{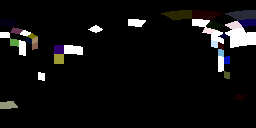}\\

\end{tabular}
\egroup
    \caption{We compare our method to existing lighting estimation methods on our real captured test set. (c) Our method correctly estimates the size and position of the light sources and produces correct shadow and specular highlights. (d) \protect{\cite{Gardner2017}} overestimates the ambient light, the position of the dominant light is off. (e) \protect{\cite{Gardner_2019_ICCV}} estimates a incorrect number of lights at incorrect locations. (f)\protect{\cite{Sato2003IFS}} results in many incorrect light sources.}
    \label{fig:ourrealcomp}
\end{figure}

%% file: fig_paulcomp.tex

\newcommand{\pualSize}{1.8cm}

\begin{figure}
   \centering
\bgroup
\tiny
\renewcommand{\tabcolsep}{0pt}
\renewcommand{\arraystretch}{0.2}
\begin{tabular}{m{\pualSize}m{\pualSize}m{\pualSize}m{\pualSize}m{\pualSize}}
\multicolumn{1}{c}{(a) Real captured} & \multicolumn{1}{c}{(b) Reference} & \multicolumn{1}{c}{(c) Gardner \protect{\cite{Gardner2017}}} & \multicolumn{1}{c}{(d) LeGendre \hspace{-0.1cm} \protect{\cite{LeGendre:2019:DLI}}} & \multicolumn{1}{c}{(e) Our result} \\

\includegraphics[width=\pualSize]{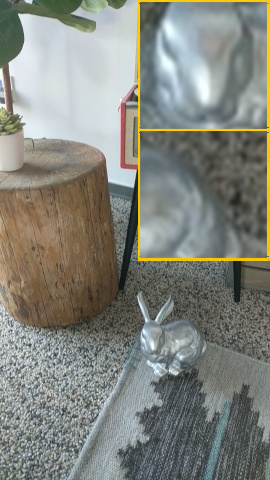}&
\includegraphics[width=\pualSize]{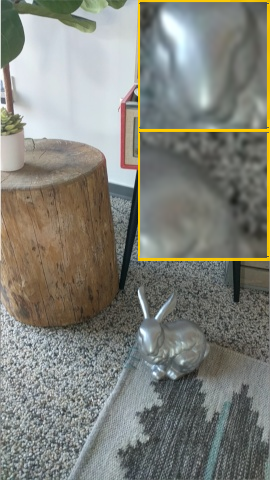}&
\includegraphics[width=\pualSize]{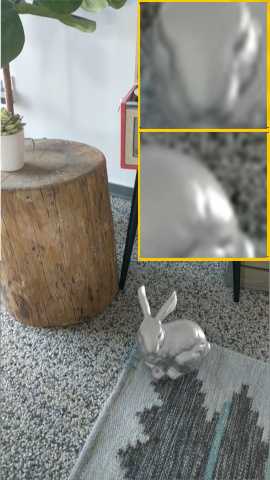}&
\includegraphics[width=\pualSize]{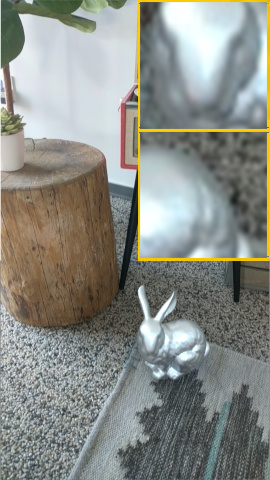}&
\includegraphics[width=\pualSize]{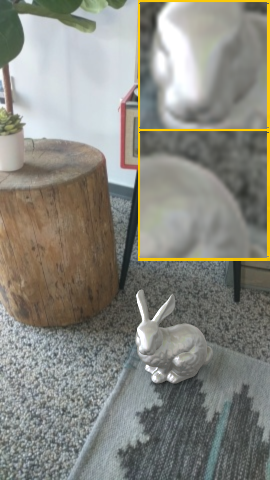}\\

\end{tabular}
\egroup
    \caption{We compare our method (with estimated depth) to existing lighting estimation methods on a real image from \protect{\cite{LeGendre:2019:DLI}} . (a) The photograph of a real 3D-printed bunny placed in the scene. Rendering results of a virtual bunny under (b) captured ground truth environment map, (c) environment map estimated by \protect{\cite{Gardner2017}}, (d) \protect{\cite{LeGendre:2019:DLI}} and (e) our method with estimated depth. Note that our result successfully reproduces specular highlight over the left side of the bunny (closed-up view inserted), similar to the ground-truth. \protect{\cite{Gardner2017}} produces wrong specular highlights on the right ear and body of the bunny, and \protect{\cite{LeGendre:2019:DLI}} results in wrong highlights all over the bunny.}
    \label{fig:paul_comp}
\end{figure}

%% file: fig_redwood.tex
\newcommand{\rwlayoutWidth}{1.7cm}

\begin{figure*}
   \centering
\bgroup
\tiny
\renewcommand{\tabcolsep}{0pt}
\renewcommand{\arraystretch}{0.2}
\begin{tabular}{m{1cm}m{\rwlayoutWidth}m{\rwlayoutWidth}m{\rwlayoutWidth} m{1cm}m{\rwlayoutWidth}m{\rwlayoutWidth}m{\rwlayoutWidth}}
\multicolumn{1}{c}{(a) Input} & \multicolumn{1}{c}{(b) Ours} & \multicolumn{1}{c}{(c) Gardner \protect{\cite{Gardner2017}}} & \multicolumn{1}{c}{(d) Gardner \protect{\cite{Gardner_2019_ICCV}}} &
\multicolumn{1}{c}{(e) Input} & \multicolumn{1}{c}{(f) Ours} & \multicolumn{1}{c}{(g) Gardner \protect{\cite{Gardner2017}}} & \multicolumn{1}{c}{(h) Gardner \protect{\cite{Gardner_2019_ICCV}}}
\\
\includegraphics[width=1cm]{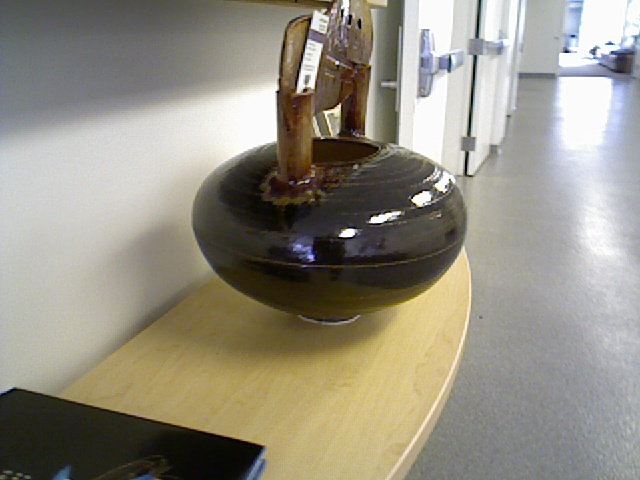}&
\includegraphics[width=\rwlayoutWidth]{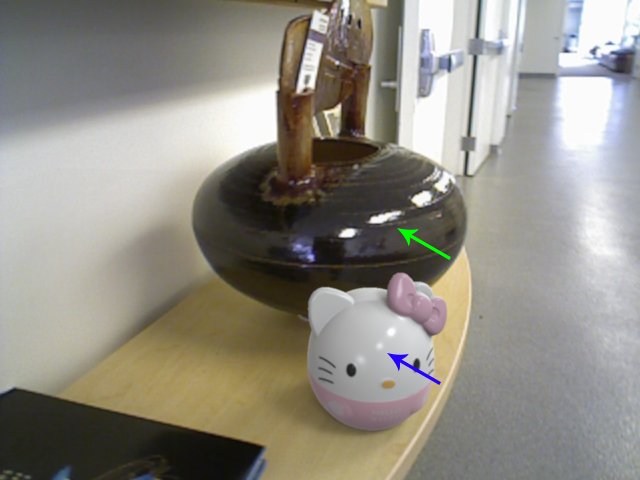}&
\includegraphics[width=\rwlayoutWidth]{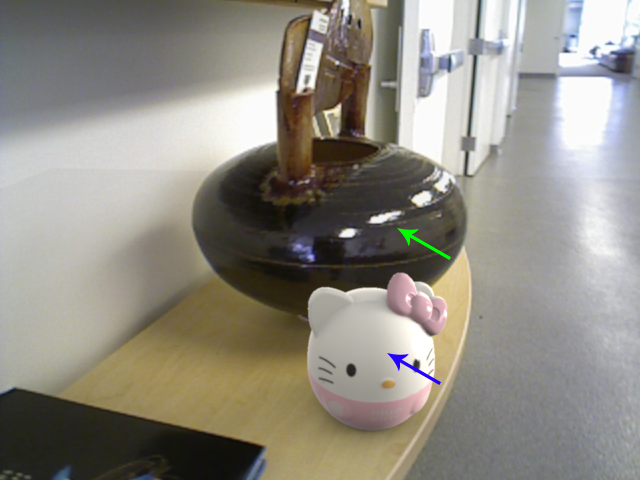}&
\includegraphics[width=\rwlayoutWidth]{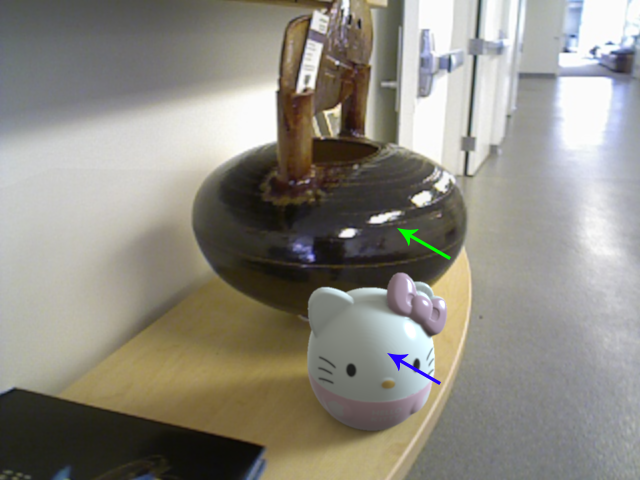}&
\includegraphics[width=1cm]{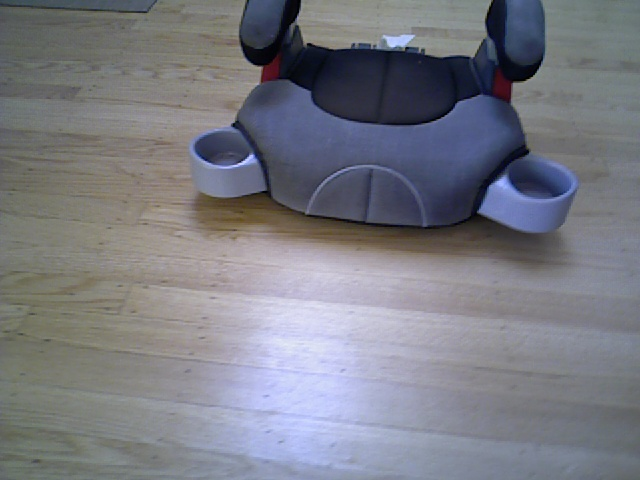}&
\includegraphics[width=\rwlayoutWidth]{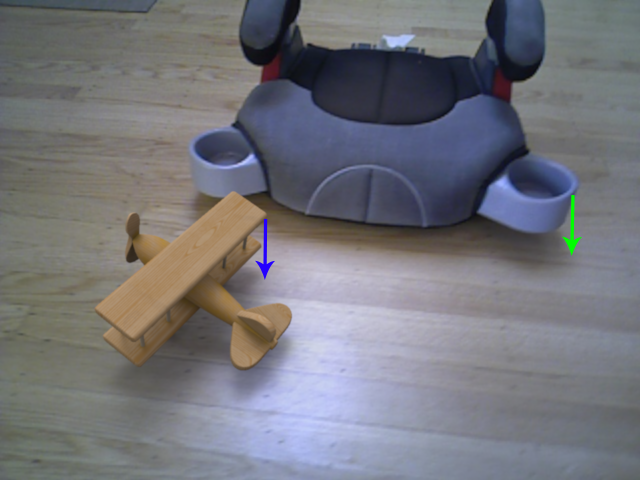}&
\includegraphics[width=\rwlayoutWidth]{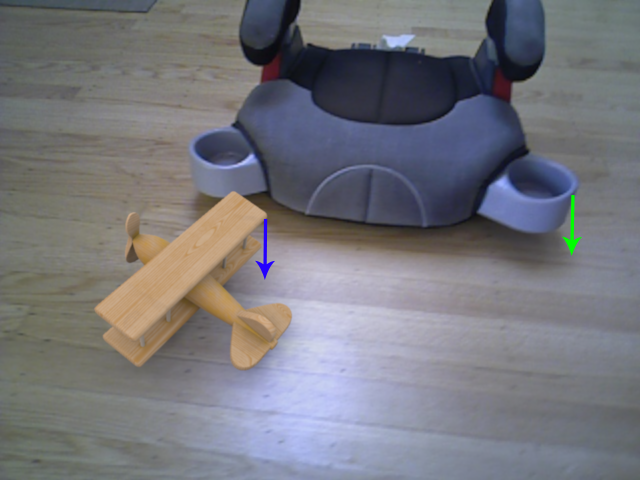}&
\includegraphics[width=\rwlayoutWidth]{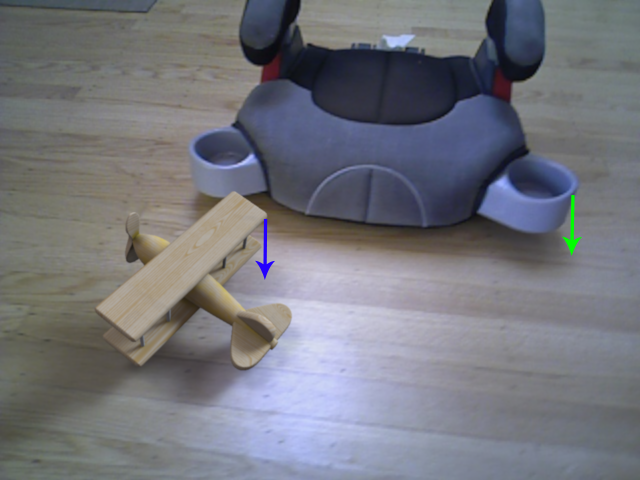}\\
&
\includegraphics[width=\rwlayoutWidth]{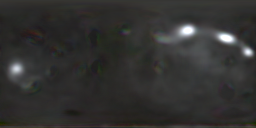}&
\includegraphics[width=\rwlayoutWidth]{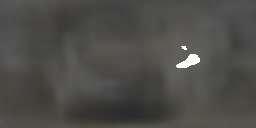}&
\includegraphics[width=\rwlayoutWidth]{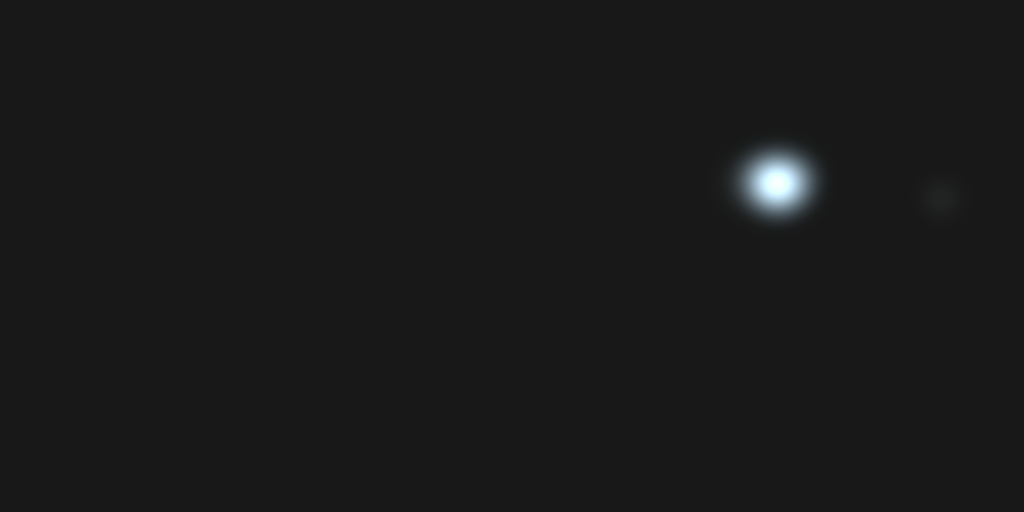}&
&
\includegraphics[width=\rwlayoutWidth]{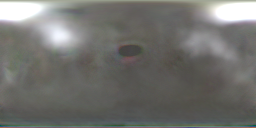}&
\includegraphics[width=\rwlayoutWidth]{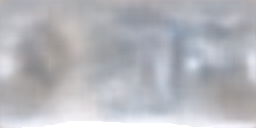}&
\includegraphics[width=\rwlayoutWidth]{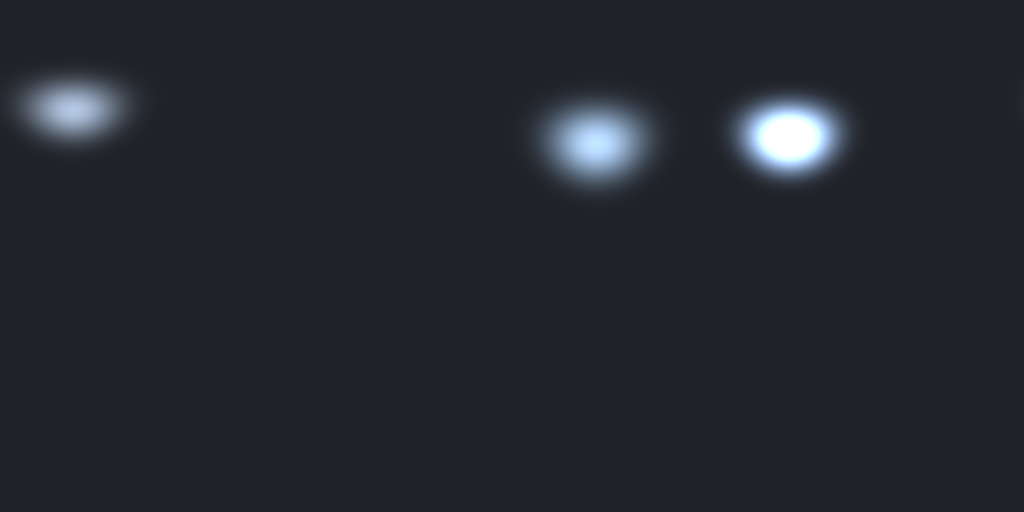}
 \\
\end{tabular}
\egroup
    \caption{Comparison on the redwood dataset \cite{Choi2016}. Left: our method faithfully reproduces the four specular highlights (blue arrow) visible from the input object (green arrow), which are missing from other methods' results. Right: our method produces shadow with direction consistent to the input (blue and green arrows). Previous methods either fail to reproduce the shadow \cite{Gardner2017} or generate shadow in the wrong direction \cite{Gardner_2019_ICCV}.}
    \label{fig:redwoodreal}
\end{figure*}

%% file: fig_ablation_net.tex

\begin{figure*}[h]
   \centering
\bgroup
\tiny 
\renewcommand{\tabcolsep}{0pt}
\renewcommand{\arraystretch}{0.2}
\begin{tabular}{m{1em} m{\fiveColumnFigWidth}m{\fiveColumnFigWidth}m{\fiveColumnFigWidth}m{\fiveColumnFigWidth}m{\fiveColumnFigWidth} c m{\fiveColumnFigWidth}m{\fiveColumnFigWidth}m{\fiveColumnFigWidth}m{\fiveColumnFigWidth}m{\fiveColumnFigWidth}}
& \multicolumn{1}{c}{Reference} & \multicolumn{1}{c}{Ours} & \multicolumn{1}{c}{without} & \multicolumn{1}{c}{without} & \multicolumn{1}{c}{Direct} & & \multicolumn{1}{c}{Reference} & \multicolumn{1}{c}{Ours} & \multicolumn{1}{c}{without} & \multicolumn{1}{c}{without} & \multicolumn{1}{c}{Direct}\\
& \multicolumn{1}{c}{} & \multicolumn{1}{c}{} & \multicolumn{1}{c}{Specular} & \multicolumn{1}{c}{Decompose} & \multicolumn{1}{c}{Regression} & & \multicolumn{1}{c}{} & \multicolumn{1}{c}{} & \multicolumn{1}{c}{Diffuse} & \multicolumn{1}{c}{Decompose} & \multicolumn{1}{c}{Regression}\\
 \rotatebox{90}{Render}&
\includegraphics[width=\fiveColumnFigWidth]{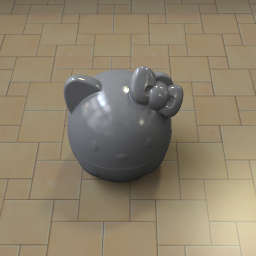}&
\includegraphics[width=\fiveColumnFigWidth]{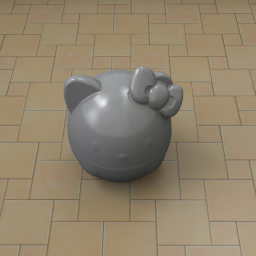}& 
\includegraphics[width=\fiveColumnFigWidth]{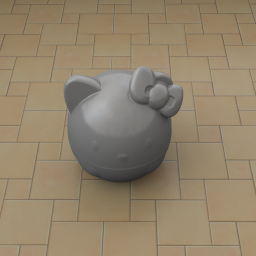}&
\includegraphics[width=\fiveColumnFigWidth]{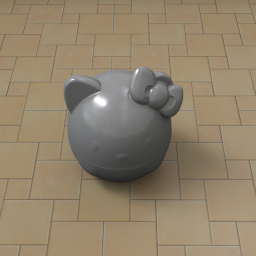}&
\includegraphics[width=\fiveColumnFigWidth]{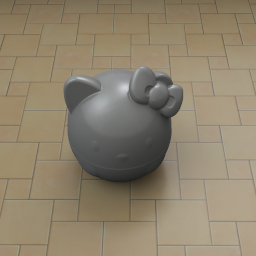}& \hspace{2pt} &
\includegraphics[width=\fiveColumnFigWidth]{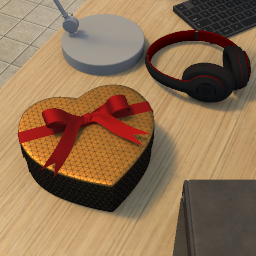}&
\includegraphics[width=\fiveColumnFigWidth]{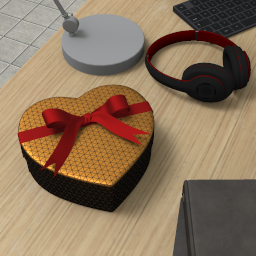}& 
\includegraphics[width=\fiveColumnFigWidth]{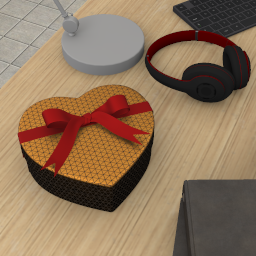}&
\includegraphics[width=\fiveColumnFigWidth]{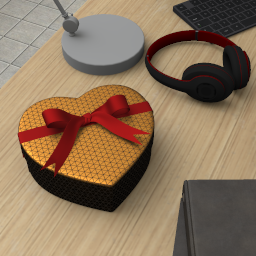}&
\includegraphics[width=\fiveColumnFigWidth]{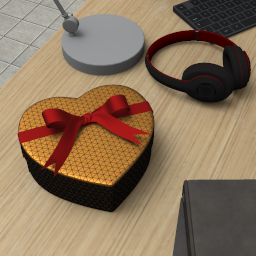}\\
 \rotatebox{90}{Light}&
\includegraphics[width=\fiveColumnFigWidth]{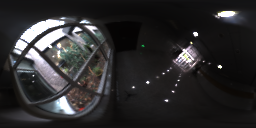}&
\includegraphics[width=\fiveColumnFigWidth]{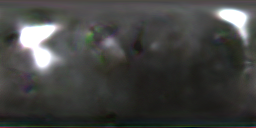}&
\includegraphics[width=\fiveColumnFigWidth]{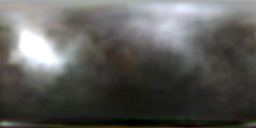}&
\includegraphics[width=\fiveColumnFigWidth]{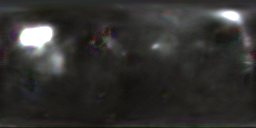} &
\includegraphics[width=\fiveColumnFigWidth]{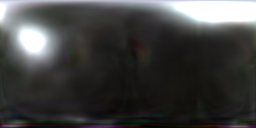}& \hspace{2pt} &
\includegraphics[width=\fiveColumnFigWidth]{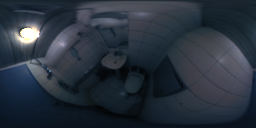}&
\includegraphics[width=\fiveColumnFigWidth]{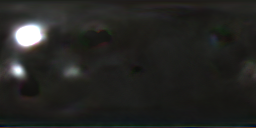}&
\includegraphics[width=\fiveColumnFigWidth]{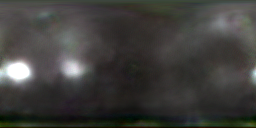}&
\includegraphics[width=\fiveColumnFigWidth]{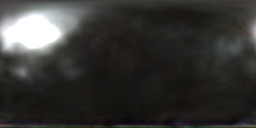}&
\includegraphics[width=\fiveColumnFigWidth]{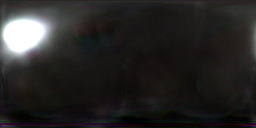}\\
\end{tabular}
\egroup
    \caption{Ablation study of our physically-based estimation. For a scene containing specular objects (left), our specular network can help to infer high-frequency lighting details from the decomposed specular reflections. For diffuse dominated scenes (right), the diffuse network plays an important role in estimating light from the diffuse shading and shadows. Without having the corresponding network or the decomposition, the system fails to produce accurate estimations. Direct regression also produces blurred light estimation. The inaccurate estimations lead to degraded rendering results, such as missing highlights (left) and shadows (right).}
    \label{fig:ablation_net} 
\end{figure*}

%% file: conclusion.tex
\section{Conclusion}
\label{sec:conclusion}

We presented a scheme for realtime environment light estimation from the RGBD appearance of individual objects, rather than from broad scene views.
By designing the neural networks to follow physical reflectance laws and infusing rendering knowledge with additional input, our method can robustly estimate environment lighting for scenes with arbitrary objects and various illumination conditions. Our recurrent convolution design also offers temporal and spatial smoothness which is critical for many AR applications. 

Although our method supports near-field illumination effects by estimating light at different local regions of the input, a potential improvement would be to estimate a near-field illumination model, combining inferences from multiple different patches, to yield a more robust solution for near-field illumination.

Currently, our method estimates environment light only based on the shading information of objects and does not require the input contains contents of the environment map. Combining scene-based method \cite{LeGendre:2019:DLI,Song_2019_CVPR,Gardner_2019_ICCV} with our object-based method would be a potential future direction, yielding better quality results.

%% file: fig_teaser.tex
\newcommand{\TeaserInputSize}{0.1}
\newcommand{\TeaserRenderSize}{0.3}
\newcommand{\TeaserCubeSize}{0.1}

\begin{teaserfigure}
  \centering
\begin{minipage}[c] {\TeaserInputSize\textwidth}
			\begin{minipage}[t]{\textwidth}
			\centering
			\includegraphics[width=\textwidth]{fig/real/01/rgb.png}
			\vspace*{-0.8cm}\caption*{Input RGB}
			\end{minipage}
			\begin{minipage}[t]{\textwidth}
			\centering
			\includegraphics[width=\textwidth]{fig/real/01/depth.png}
			\vspace*{-0.8cm}\caption*{Input Depth}
			\end{minipage}
		\end{minipage}
		\begin{minipage}[c] {\TeaserRenderSize\textwidth}
			\centering
			\includegraphics[width=\textwidth]{fig/real/01/Result_Pred.png}
			\vspace*{-0.8cm}\caption*{Virtual object rendered into the input image}
		\end{minipage}
		\begin{minipage}[c] {\TeaserCubeSize\textwidth}
			\begin{minipage}[t]{\textwidth}
			\centering
			\includegraphics[width=\textwidth]{fig/real/01/Result_Pred_light_view.png}
			\vspace*{-0.8cm}\caption*{Left result}
			\end{minipage}
			\begin{minipage}[t]{\textwidth}
			\centering
			\includegraphics[width=\textwidth]{fig/real/04/Result_Pred_light_view.png}
			\vspace*{-0.8cm}\caption*{Right result}
			\end{minipage}
		\end{minipage}
		\begin{minipage}[c] {\TeaserInputSize\textwidth}
			\begin{minipage}[t]{\textwidth}
			\centering
			\includegraphics[width=\textwidth]{fig/real/04/rgb.png}
			\vspace*{-0.8cm}\caption*{Input RGB}
			\end{minipage}
			\begin{minipage}[t]{\textwidth}
			\centering
			\includegraphics[width=\textwidth]{fig/real/04/depth.png}
			\vspace*{-0.8cm}\caption*{Input Depth}
			\end{minipage}
		\end{minipage}
		\begin{minipage}[c] {\TeaserRenderSize\textwidth}
			\centering
			\includegraphics[width=\textwidth]{fig/real/04/Result_Pred.png}
			\vspace*{-0.8cm}\caption*{Virtual object rendered into the input image}
		\end{minipage}
		\vspace*{-0.5cm}
  \caption{From an RGBD image crop of a 3D object and its local area, our method automatically estimates the all-frequency environment lighting at the object location.
   With the estimated lighting, virtual 3D objects (toy radio and potted cactus) can be rendered into the input image with consistent shading and shadow effects.}
  \label{fig:teaser}
\end{teaserfigure}